\documentclass[11pt]{article}

\usepackage[margin=1in]{geometry}
\usepackage[round,authoryear]{natbib}
\usepackage[hidelinks]{hyperref}
\usepackage{microtype}
\usepackage{placeins} 
\usepackage{flafter}  

\usepackage{amsmath}
\usepackage{amsfonts}
\usepackage{algorithmicx}
\usepackage[noend]{algpseudocode}
\usepackage{algorithm}
\usepackage{array}
\usepackage{subfig}
\usepackage{textcomp}
\usepackage{url}
\usepackage{verbatim}
\usepackage{graphicx}
\usepackage{xcolor}
\usepackage{xspace}
\usepackage{booktabs}
\usepackage{tabularx}

\newcommand{\IOHexp}{\textsc{{IOHexperimenter}}\xspace}
\DeclareMathOperator{\tr}{tr}

\newcommand{\pkg}[1]{{\selectfont #1}}

\newcommand{\modcma}{\pkg{ModCMA}\xspace}
 
\newcommand*\Let[2]{\State #1 $\gets$ #2}

\newcommand{\z}{\ensuremath{\mathbf{z}}\xspace}
\newcommand{\x}{\ensuremath{\mathbf{x}}\xspace}
\newcommand{\y}{\ensuremath{\mathbf{y}}\xspace}
\newcommand{\m}{\ensuremath{\mathbf{m}}\xspace}
\newcommand{\C}{\ensuremath{\mathbf{C}}\xspace}
\newcommand{\A}{\ensuremath{\mathbf{A}}\xspace}
\newcommand{\B}{\ensuremath{\mathbf{B}}\xspace}
\newcommand{\D}{\ensuremath{\mathbf{D}}\xspace}
\newcommand{\pc}{\ensuremath{\mathbf{p}_c}\xspace}
\newcommand{\lb}{\ensuremath{\mathbf{lb}}\xspace}
\newcommand{\ub}{\ensuremath{\mathbf{ub}}\xspace}
\newcommand{\ps}{\ensuremath{\mathbf{p}_\sigma}\xspace}

\newcommand{\normd}{\ensuremath{\mathcal{N}(\mathbf{0}, \mathbf{I})}\xspace}

\newcommand{\cmac}{($\mu/\mu_w$, $\lambda$)-CMA-ES\xspace}
\newcommand{\g}{\ensuremath{^{(g)}}}
\renewcommand{\ng}{\ensuremath{^{(g+1)}}}
\newcommand{\norm}[1]{\left\lVert#1\right\rVert}
\newcommand{\diag}[1]{\ensuremath{\operatorname{diag}(#1)}}

\newcommand{\blockheader}[2]{%
  \Statex\vspace{0.01em}
  \Statex \colorbox{#1}{\parbox{\dimexpr\linewidth-2\fboxsep}{\textbf{#2}}}
}

\begin{document}

\title{The Modular CMA-ES: A Framework for Modern Evolution Strategies}

\author{%
Jacob de Nobel \and
Diederick Vermetten \and
Anna V. Kononova \and
Carola Doerr \and
Thomas B\"ack
}
\date{}

\maketitle

\begin{abstract}

Since their introduction, modern evolution strategies, such as the Covariance Matrix Adaptation Evolution Strategy (CMA-ES), have become established as powerful methods for continuous black-box optimization. This success has led to a wide range of proposed modifications, each designed to improve performance or behavior in specific optimization scenarios. However, because these developments have largely been introduced and studied in isolation, their interactions remain comparatively underexplored.
In this paper, we present the Modular CMA-ES (\modcma), a configurable framework that integrates a wide range of mechanisms from modern evolution strategies within a single implementation. By decomposing CMA-ES into modules with interchangeable options for sampling, selection and recombination, step-size adaptation, matrix adaptation, and restarting, \modcma enables systematic exploration of a large design space of modern evolution strategies and facilitates the construction, comparison, and automated configuration of new algorithm variants.
We illustrate the benefits of this modular approach through two example studies. First, we compare several matrix-adaptation mechanisms in terms of their computational cost and optimization performance. Second, we use automated algorithm configuration to specialize \modcma to individual benchmark problems and analyze the resulting configurations. Together, these examples demonstrate how the framework can be used both to study individual algorithmic design choices and to explore their combinations in a systematic and reproducible manner.
\end{abstract}

\begin{center}
\end{center}

\section{Introduction}

Many real-world optimization problems can only be accessed as black boxes: the objective function is expensive, noisy, or lacks analytical structure, and gradient information is unavailable or unreliable. Even when the problem is not purely black-box in nature, methods that treat it as such can be effective, e.g., when problem knowledge is not linked to existing solution techniques, or when the effort required to integrate problem knowledge into a custom solver is too large. Evolution strategies (ES) \citep{Rechenberg73, schwefel1965kybernetische} provide a robust class of methods for such settings, with the Covariance Matrix Adaptation Evolution Strategy (CMA-ES)~\citep{hansen2001completely} having emerged as one of the most widely used algorithms for continuous black-box optimization. CMA-ES combines strong empirical performance on benchmarks \citep{hansen2009benchmarking}, a comparatively small set of sensitive hyperparameters, and strong invariance properties.

Over the years, numerous CMA-ES variants and extensions have been proposed. These include restart strategies for multimodal functions~\citep{auger2005restart,bipop}, alternative sampling schemes such as mirrored and orthogonal sampling~\citep{brockhoff_mirrored_2010,wang_mirrored_2014}, different step-size adaptation mechanisms~\citep{hansen_cma-es_2008, msr, qualitative_ssa_krause_glas_2017}, modifications of the covariance update~\citep{ros2008simple,suttorp2009efficient,beyer_simplify_2017}, and specialized mechanisms for dealing with boundary constraints \citep{Dufosse21augmented}. Each of these contributions is typically introduced and evaluated in isolation, often with its own implementation and experimental protocol \citep{van_rijn_evolving_2016}.

This fragmented development has two consequences. First, it is difficult to understand how different algorithmic components interact \citep{van_rijn2017alg}. For instance, the impact of a given sampling strategy may depend critically on the choice of matrix adaptation scheme or restart strategy. Second, practitioners face a practical barrier: selecting a suitable CMA-ES variant for a given application requires navigating a growing collection of specialized implementations, each with its own configuration interface and assumptions \citep{NikolikjMTE25}. As a result, many powerful ideas remain underexplored in combination, and advanced variants are less accessible to non-expert users.

A natural response to this situation is to view algorithmic design itself as a compositional problem. Earlier work on Modular Evolutionary Algorithms (ModEA)~\citep{van_rijn_evolving_2016} and the first version of the modular CMA-ES~\citep{denobel2021tuning} followed exactly this idea: they decomposed the algorithm into a set of modules, such as sampling, selection, step-size adaptation, matrix adaptation, and restarting, with multiple interchangeable options exposed through a configurable design space. This modularization enables systematic studies of how individual components influence performance~\citep{denobel2021tuning,nikolikj2024quantifying}, facilitates automated algorithm configuration and dynamic algorithm selection~\citep{Vermetten20CASH,vermetten2023switch,thomaser2023optimizing,long2024landscape,thomson2025xai}, and lowers the barrier for implementing and testing new CMA-ES variants. 

The Modular CMA-ES has already been used to develop and evaluate new algorithmic ideas~\citep{dinu2025adaptive,denobel2024avoiding,denobel2024ecta,bonet2023performance,denobel2025abnormal}, and it is featured in several benchmarking studies~\citep{eaf,vermetten2023ma,vanStein2025Explainable,vermetten2024large,vermetten2023analysis,nikolikj2025benchmarking, kononova2025xai}. Moreover, the framework has been employed in studies of algorithm behaviour~\citep{kostovska2022per,jankovic2022trajectory,vermetten2022analyzing, van2024deepbias, Vermetten2022structural} and in landscape-informed tuning and configuration~\citep{kostovska2022importance,kostovska2023using,nikolikj2022identifying,kostovska2024using}. The practical value of treating the CMA-ES as a modular algorithm is demonstrated by its use in real-world optimization tasks, such as biological modeling~\citep{denobel2024Biophysics}, mechanical engineering~\citep{long2024generating,thomaser2023transfer,long2025surrogate,thomaser2023real} and optics~\citep{patel2025design}.

In this paper, we present the current incarnation of the \emph{Modular CMA-ES}, denoted \modcma, a substantially extended and reengineered framework building on these earlier efforts~\citep{van_rijn_evolving_2016, denobel2021tuning}. \modcma provides a unified implementation of a wide range of CMA-ES variants within a single, component-based architecture. The framework cleanly separates the algorithm into coherent modules and offers multiple options for each, resulting in fourteen implemented modules (Table~\ref{modcma:tab:options}) and over 14 million possible CMA-ES instantiations. The core is reimplemented in C++ using \textsc{Eigen}~\citep{eigen} for efficient linear algebra, with a Python interface for convenient experimentation.

Beyond describing the framework itself, we use \modcma as a vehicle to study aspects of modular algorithm design and configuration. In particular, we compare different matrix adaptation mechanisms in terms of computational cost and optimization performance, and we show how automated configuration methods can exploit modularity to tailor CMA-ES variants to specific problems.

Our main contributions are as follows:
\begin{enumerate}
    \item We introduce and document the current version of \modcma, a modular CMA-ES framework integrating fourteen interchangeable modules covering sampling, selection, step-size adaptation, matrix adaptation, and restarting. The framework unifies many CMA-ES variants from the literature within a single, coherent implementation.
     \item We provide a systematic empirical comparison of matrix adaptation mechanisms, quantifying their computational trade-offs and optimization performance across the BBOB benchmark suite \citep{hansen2020coco}. We show that efficient alternatives such as Cholesky- and matrix-based updates can match, and in some cases surpass, the standard covariance update while offering substantial runtime reductions. 
    \item We demonstrate how \modcma enables automated algorithm configuration by conducting an illustrative tuning experiment with \textsc{SMAC} \cite{smac} on individual BBOB functions and by comparing the resulting specialized CMA-ES variants to established algorithms from the COCO archives~\cite{coco_bbob_data_archive}. 
    \item We use the tuning data as a case study to show how \modcma supports explainable analysis of algorithm configuration results. We then apply \textsc{SHAP}~\citep{shap} to identify how module choices contribute to the performance of tuned configurations~\citep{van2024explainable}.
    \item We make \modcma freely available as an open-source package under an MIT license at \url{https://github.com/IOHprofiler/ModularCMAES}. 
\end{enumerate}

The remainder of the paper is organized as follows. Section~\ref{modcma:sec:overview} introduces the modular decomposition of CMA-ES and provides an overview of the available modules and options. Sections~\ref{modcma:sec:sampling}--\ref{sec:mod:other} describe the components of the framework in detail.
Section~\ref{modcma:sec:matrix:exp} zooms in on the matrix adaptation module, presenting runtime measurements and benchmarking experiments. Section~\ref{modcma:sec:hyper} reports on the tuning study and the associated module analysis. Section~\ref{sec:conclusion} concludes with a discussion of the implications and directions for future work. 

To ensure reproducibility, we made all code and data used throughout this paper, as well as additional figures, available in our Zenodo repository~\cite{zenodo}.

\section{Overview}
\label{modcma:sec:overview}

\subsection{The CMA-ES Algorithm}
\label{pre:sec:cma}
The CMA-ES is a stochastic, population-based algorithm for continuous black-box optimization, originally introduced in~\citep{hansen2001completely}. At its core, CMA-ES maintains a multivariate normal search distribution $\mathcal{N}(\m, \sigma^2\C)$ parameterized by a mean vector \m, a global step size $\sigma$, and a covariance matrix \C. In each generation, a population of  $\lambda$ candidate solutions is sampled from this distribution, evaluated, and recombined to update the distribution parameters. CMA-ES derives much of its effectiveness from its adaptation mechanisms. Step-size adaptation regulates the global scale of mutations by controlling $\sigma$, while covariance matrix adaptation gradually reshapes the sampling distribution to align with directions of successful search steps \citep{hansen2016cmatut}. Together, these mechanisms confer strong invariance properties under affine transformations of the search space and make CMA-ES well-suited to solve ill-conditioned, non-separable optimization problems \citep{hansen1996adapting}.

\begin{algorithm}[!htbp] 
\small
  \caption{Simplified pseudocode for the \cmac. The all-zero vector of length $d$ is denoted by $\mathbf{0}^d$. The generation counter $g$ is omitted for clarity wherever possible. The algorithm is divided into five functional components, each separated by a colored header.}
  \label{alg:modcma}
  \begin{algorithmic}[1]
    \Require{Initial step size $\sigma_0$, $d \in \mathbb{N}_+$, initial point $\x_0 \in \mathbb{R}^d$, population size $\lambda_0 \geq 4$, and recombination weights $\mathbf{w}\in \mathbb{R}^\mu$. Constants: $\mu_{\text{eff}}, c_c, c_m, d_\sigma, c_\sigma, c_1, c_\mu$ set according to~\cite{hansen2016cmatut}.}
    \Statex
    \Procedure{CMA-ES}{} 
    \While{termination criteria not met} 
    \blockheader{red!15}{\hspace{10pt}Restarting}
            \State $\sigma \gets \sigma_0,\ \mu \gets \lfloor\frac{\lambda_0}{2}\rfloor, \m \gets \x_0$ \Comment{Potentially modify $\sigma_0, \lambda_0, \x_0$}
            \State $\C \gets \mathbf{I},\ \B \gets \mathbf{I},\ \D \gets \mathbf{I},\ \A \gets \mathbf{I}$
            \State $\pc \gets \mathbf{0}^d,\ \ps \gets \mathbf{0}^d$ 
            
            \Repeat
            \blockheader{blue!15}{\hspace{25pt}Sampling}
            \For{$i \gets 1 \textrm{ to } \lambda$}  
                \State $\z_{i} \sim \mathcal{N}(\mathbf{0}, \mathbf{I})$ 
                \Let{$\y_{i}$}{$\A \z_{i}$} 
                \Let{$\x_{i}$}{$\m + \sigma \y_{i}$}
            \EndFor

            \blockheader{orange!20}{\hspace{25pt}Selection and Recombination}
            \Let{$\m\ng$}{$\m\g + \sum_{i=0}^{\mu} w_i (\x_{i:\lambda} - \m)$} \Comment{$\x_i$ sorted w.r.t. $f(\x_i)$}\label{alg:line:modmcasel}
            \Let{$\langle\y\rangle_w$}{$(\m\ng - \m\g) / \sigma$}

            \blockheader{teal!20}{\hspace{25pt}Step-size Adaptation (SSA)}
            \Let{$\ps$}{$(1 - c_\sigma)\ps + \sqrt{c_\sigma(2-c_\sigma) \mu_{\text{eff}}} \C^{-\frac{1}{2}} \langle\y\rangle_w$}
            \Let{$\sigma$}{$\sigma \cdot \exp \left(\frac{c_\sigma}{d_\sigma} \left( \frac{\|\ps\|}{\mathbb{E}\|\mathcal{N}(\mathbf{0}, \mathbf{I})\|} - 1 \right)\right)$}

            \blockheader{purple!15}{\hspace{25pt}Covariance Matrix Adaptation (CMA)}
            \Let{$\pc$}{$(1 - c_c)\pc + \sqrt{c_c(2-c_c) \mu_{\text{eff}}} \langle\y\rangle_w$}  \Comment{$h_\sigma$ is omitted for simplicity.}
            \State $\C \gets (1 -c_1 - c_\mu \sum_{i=0}^{\mu} w_i)\ \C +  c_1 \pc( \pc)^\top +  c_\mu \sum_{i=0}^{\mu} w_i \y_{i:\lambda} (\y_{i:\lambda})^\top$
            \State $\B \D^2\B^\top \gets \C$ \Comment{Decompose \C}
            \Let{$\C^{-\frac{1}{2}}$}{$\B\D^{-1}\B^\top$}
            \Let{$\A$}{$\B\D$} 
            \Until{restart triggered} 
        \EndWhile
    \EndProcedure
  \end{algorithmic}
\end{algorithm}

In its canonical form, the \cmac follows a generation-based loop consisting of sampling, selection and recombination, followed by parameter control, optionally augmented with restart strategies to handle multimodality. Algorithm~\ref{alg:modcma} presents a simplified pseudocode description. Certain specifics, such as the exact computation of the recombination weights $\mathbf{w}$, the learning rates, and the parameter $h_\sigma$ that is used to stall the update of the covariance evolution path when the step size becomes too small, are omitted for clarity. The interested reader is referred to the tutorial by~\citet{hansen2016cmatut} for implementation details. 

The pseudocode is structured around five functionally distinct components: restarting, sampling, selection and recombination, step-size adaptation, and covariance (or matrix) adaptation. These components are highlighted by colored blocks in Algorithm~\ref{alg:modcma}. Importantly, while these components interact through shared state variables, they are conceptually separable and admit multiple alternative implementations, facilitating a modular approach. 

\subsection{From CMA-ES to Modular CMA-ES}

The Modular CMA-ES (\modcma) is built on the observation that CMA-ES naturally decomposes into a small number of coherent, interchangeable \emph{components}, such as sampling, selection and recombination, step-size adaptation, matrix adaptation, and restarting. \modcma exposes the algorithmic choices within these components as explicit \emph{modules}, each of which can be independently configured. This design follows the Modular Evolutionary Algorithms (ModEA) paradigm introduced by \citet{van_rijn_evolving_2016} and extends an earlier modular CMA-ES implementation~\citep{denobel2021tuning}.

Following the structure shown in Algorithm~\ref{alg:modcma}, each component is represented by one or more modules, with several alternative \emph{options} corresponding to mechanisms from the literature. For example, the \emph{Matrix Adaptation} module includes options such as \texttt{Covariance}, \texttt{Separable}, and \texttt{Cholesky}. We use the term \emph{configuration} for a complete specification of the module options together with the relevant numerical parameter settings. When referring only to different combinations of module options, with numerical parameters kept at their prescribed defaults, we use the term \emph{modular CMA-ES variant}.

\begin{table}[!htbp]
\centering
\small
\caption{Implemented modules and configuration options available in \modcma. The default option is provided in boldface. The component affected by the algorithm is shown alongside each module.
}
\label{modcma:tab:options}
\begin{tabularx}{\linewidth}{@{} l l l X @{}}
\toprule
\textbf{Module} & \textbf{Component} & \textbf{\#} & \textbf{Available Options} \\
\midrule
\emph{Base Sampler} & Sampling & 3 & \textbf{\texttt{Uniform}}, \texttt{Sobol}, \texttt{Halton} \\
\emph{Sample Transformation} & Sampling & 6 & \textbf{\texttt{Gaussian}}, \texttt{Uniform}, \texttt{Laplace}, \texttt{Logistic}, \texttt{Cauchy}, \texttt{Double Weibull} \\
\emph{Mirrored Sampling} & Sampling & 3 & \textbf{\texttt{None}}, \texttt{Mirrored}, \texttt{Pairwise} \\
\emph{Orthogonal Sampling} & Sampling & 2 & \textbf{\texttt{Disabled}}, \texttt{Enabled} \\
\emph{Threshold Convergence} & Sampling & 2 & \textbf{\texttt{Disabled}}, \texttt{Enabled} \\
\emph{Boundary Correction} & Sampling & 7 & \textbf{\texttt{None}}, \texttt{Resample}, \texttt{Mirror}, \texttt{COTN}, \texttt{Saturate}, \texttt{Uniform Resample}, \texttt{Toroidal} \\
\emph{Elitism} & Sel. \& Rec. & 2 & \textbf{\texttt{Disabled}}, \texttt{Enabled} \\
\emph{Sequential Selection} & Sel. \& Rec. & 2 & \textbf{\texttt{Disabled}}, \texttt{Enabled} \\
\emph{Recombination Weights} & Sel. \& Rec. & 3 & \textbf{\texttt{Default}}, \texttt{Equal}, \texttt{Exponential} \\
\emph{Step-Size Adaptation} & SSA & 9 & \textbf{\texttt{CSA}}, \texttt{TPA}, \texttt{MSR}, \texttt{xNES}, \texttt{m-xNES}, \texttt{lp-xNES}, \texttt{PSR}, \texttt{SR}, \texttt{SA} \\
\emph{Active Update} & CMA & 2 & \textbf{\texttt{Disabled}}, \texttt{Enabled} \\
\emph{Matrix Adaptation} & CMA & 7 & \textbf{\texttt{Covariance}}, \texttt{None}, \texttt{Separable}, \texttt{Matrix}, \texttt{Cholesky}, \texttt{CMSA}, \texttt{Natural Gradient} \\
\emph{Repelling} & Restarting & 2 & \textbf{\texttt{Disabled}}, \texttt{Enabled} \\
\emph{Restart Strategy} & Restarting & 4 & \textbf{\texttt{None}}, \texttt{Restart}, \texttt{IPOP}, \texttt{BIPOP} \\
\bottomrule
\end{tabularx}
\end{table}

\subsection{A Modular Structure}

In total, \modcma implements fourteen modules, each affecting a specific component of the algorithm. These include alternative step-size adaptation mechanisms~\citep{denobel2021tuning}, flexible sampling distributions~\citep{denobel2025abnormal}, a repelling restart mechanism for multimodal optimization~\citep{denobel2024avoiding}, and several matrix adaptation mechanisms that allow variants such as MA-ES~\citep{beyer_simplify_2017} and separable CMA-ES~\citep{ros2008simple} to be instantiated within the same framework. Table~\ref{modcma:tab:options} summarizes the modules and their available options, grouped by the algorithmic components introduced in Algorithm~\ref{alg:modcma}.

Not every combination of module options constitutes a distinct algorithm. In particular, \emph{Active Update} has no effect when \emph{Matrix Adaptation} is set to \texttt{None}, since no matrix update is performed. Similarly, \emph{Repelling} is inactive when the \emph{Restart Strategy} is set to \texttt{None}, as the mechanism operates across restarts. Accounting for these dependencies, the fourteen modules define
$$
3 \cdot 6 \cdot 3 \cdot 2 \cdot 2 \cdot 7 \cdot 2 \cdot 2
\cdot 3 \cdot 9 \cdot 13 \cdot 7
= 14{,}859{,}936
$$
distinct modular CMA-ES variants when numerical parameters are kept at their prescribed defaults. The subsequent sections describe each module and its options in detail, organized by the affected algorithmic component.

\section{Sampling}
\label{modcma:sec:sampling}
As shown in lines 7 to 10 in Algorithm \ref{alg:modcma}, the sampling procedure of the CMA-ES follows three steps; 1) a sampling of new points from a multivariate normal distribution, i.e. $\z_i \sim \normd$, which are then 2) transformed by $\A$ and 3) scaled by the mean \m and global step size $\sigma$. Practically, each component $z_{i, j}$ of the mutation vector $\z_i$ is generated independently. This can be done by  \emph{inverse transformation sampling}: first we generate a number $u_j \sim \mathcal{U}(0, 1]$, and then use the percent-point function (PPF) of the Gaussian distribution $Q_{\text{Gauss}}(p)$ to generate the required standard normal variable: $z_{i, j} = Q_{\text{Gauss}}(u_j) \sim \mathcal{N}(0, 1)$. In the Modular CMA-ES, five modules have been implemented that operate on the sampling component, explained in the next paragraphs. 

\subsection{Base Sampler} 
In standard random sampling, the random uniform numbers that are used to construct the mutation vectors are generated using a pseudo-random number generator (PRNG). Specifically, a Mersenne-Twister general-purpose PRNG provides a source of (deterministic) randomness for generating uniformly distributed numbers. However, especially for smaller samples, random numbers generated using this method can be distributed very unevenly \citep{halton1960efficiency}. Instead, \emph{Low-Discrepancy Sequences} provide sequences of pseudo-randomly generated points that more evenly cover the domain \citep{clement2023constructing}. In ES, quasi-random point sets have been used as an alternative sampling strategy \cite{auger2005quasi_random,teytaud2007dcma, denobel2024ecta}, replacing the source of randomness of $u_j$. In \modcma, both \texttt{Halton} \citep{halton1960efficiency} and \texttt{Sobol} \citep{sobol1967distribution} sequences can be used instead of the standard PRNG.

\subsection{Sample Transformation}
Instead of transforming $u_j$ to a normally distributed random sampling using the Gaussian PPF, we can transform $u_j$ to another continuous multivariate distribution. While several requirements apply to the distribution \citep{rudolph1994evolutionary,Beyer01}, six functional distributions have been implemented (see Table \ref{modcma:tab:options}). This is explained in more detail by \citet{denobel2025abnormal}, which demonstrates a potential use case for non-Gaussian mutations in CMA-ES. By default, the maximum-entropy \citep{rudolph1994evolutionary} \texttt{Gaussian} distribution is used. 

\subsection{Mirrored Sampling} 
For every newly sampled point, its mirror image is added to the population by reversing its sign \cite{brockhoff_mirrored_2010}. Practically, this means only $\lceil\frac{\lambda}{2}\rceil$ new random samples are generated for every generation, and that after the $i$-th sample $\z_i$ is drawn randomly, sample $\z_{i+1} = -\z_i$. Mirrored sampling, when using weighted recombination and CSA, can introduce a bias toward smaller step sizes, potentially causing premature convergence \cite{auger_mirrored_2011}. \textit{Pairwise Selection} aims to counteract this by using only the best point (based on $f(\x)$) of each mirrored pair in the parameter update. 

\subsection{Orthogonal Sampling}
Orthogonal sampling extends the idea of mirrored sampling by ensuring that all (randomly) sampled mutation vectors are orthogonal using a Gram-Schmidt procedure \citep{bjorck1994numerics}. This concept was introduced in \citep{wang_mirrored_2014}, combining orthogonalization with mirrored sampling and pairwise selection. This extension results in more evenly distributed offspring, providing better local coverage of the search space. This was demonstrated in \citet{wang_mirrored_2014}, where orthogonal sampling improved local convergence and performance on the BBOB benchmark compared with the mirrored sampling method.

\subsection{Threshold Convergence}
\emph{Threshold Convergence} aims to improve exploration in ES, particularly effective for complex multimodal optimization problems \citep{piad-morffis_evolution_2015}. The method imposes a minimum distance between parent and offspring solutions to prevent premature convergence. It introduces a threshold $T\g$ that decays based on the number of generations $g$ and enforces a minimal distance a sampled point $\x_i$ must have from the population mean $\m$. If $\norm{\x_i - \m} < T\g$ the sampled $\x_i$ is `pushed' farther from $\m$, by reflecting it across the threshold radius.

\subsection{Boundary Correction}
If candidate solutions are sampled outside the search domain, they can be transformed back into it by applying a boundary-correction operation. Specifically, if a point $\x$ is sampled outside the (box-constrained) search space bounded by $[\lb, \ub]$, where $\lb \in \mathbb{R}^d$ and $\ub \in \mathbb{R}^d$denote the lower and upper bounds of the domain, then an operation can be applied to `repair' the solutions' violating coordinates. Seven options for performing boundary correction have been implemented in \modcma, based on the methods described in \citet{DBLP:journals/isci/CaraffiniKC19}:

\begin{enumerate}
    \item \texttt{None}: No correction is applied to infeasible coordinates of solutions. This may lead to undesirable behavior if $f(\x)$ is undefined outside the bounds. 
    \item \texttt{Resample}: If a solution $\x_i$ is out of bounds (oob), resample $\z_i$ until $\x_i$ is within the bounds. This may lead to stagnation, particularly when $\m$ is close to the boundary. 
    \item \texttt{Mirror} Correction Strategy: Mirrors all infeasible coordinates of a solution with respect to its closest boundary. For example, if $x_{i, j} > ub_{i, j}$ then $x_{i, j} = ub_{i, j} - (x_{i, j} - ub_{i, j})$.
    \item \texttt{Uniform Resample}: Replaces all infeasible coordinates of a solution with new coordinates sampled uniformly at random within the search space. 
    \item Complete One-tailed Normal Correction Strategy: All infeasible coordinates are replaced with new coordinates inside the search space according to a rescaled one-sided normal distribution centered on the boundary. The standard deviation of this distribution equals $\frac{\sigma\g}{3}$ where $\sigma\g$ is the global step-size. 
    \item Saturation Correction Strategy: All infeasible coordinates are set to the closest corresponding bound, e.g., when $x_{i, j} > ub_{i, j}$ then $x_{i, j} = ub_{i, j}$.  
    \item \texttt{Toroidal} Correction Strategy: All infeasible coordinates get reflected off the opposite boundary. So, when $x_{i, j} > ub_{i, j}$ then $x_{i, j} = lb_{i, j}$.   
\end{enumerate}

Note that applying boundary correction effectively changes the solution \emph{after} sampling. To correctly use these altered points in the parameter update, we must repair the solutions' intermediate values accordingly. After boundary correction, for every `repaired' point $\x_i'$, the values of $\y_i'$, and $\z_i'$ are also calculated as follows:
\begin{align}\label{modcma:eq:repair}
    \x_i' &= \textsc{Repair}(\x_i) \\
    \y_i' &= (\x_i' - \m\g) / \sigma\g \\
    \z_i' &= \A^{-1}\y_i'
\end{align}

\section{Selection and Recombination}
\label{modcma:sec:selection}
For selection and recombination, three modules are implemented. In general, the CMA-ES generates $\lambda$ offspring every generation, and uses the $\mu$ best individuals to perform weighted intermediate recombination (see line~\ref{alg:line:modmcasel} in Algorithm \ref{alg:modcma}).  

\subsection{Elitism}
The elitism module determines whether CMA-ES uses $(\mu + \lambda)$ selection instead of the default $(\mu, \lambda)$ scheme. In $(\mu + \lambda)$ selection, both the $\mu$ parent individuals, i.e., the $\mu$ selected individuals from the previous generation, and the newly generated $\lambda$ offspring are considered during selection~\citep{igel2007covariance}. Elitism ensures that the best solution found so far is preserved, thereby improving local convergence. However, in real-valued evolution strategies, non-elitist selection is generally preferred because it exhibits stronger exploratory behavior \citep{schwefel1987collective, Beyer01}. Additionally, solutions sampled in a previous generation must be appropriately transformed for the parameter update to work. This is achieved via the same procedure as with boundary correction, where the values $\y_i'$ and $\z_i'$ are recomputed from an elitist $\x_i$ (see Equation \ref{modcma:eq:repair}).

\subsection{Recombination Weights}  
This module determines how the recombination weights $\mathbf{w} \in \mathbb{R}^\mu$ are generated. The default option generates $\mathbf{w}$ according to the procedure defined in \citet{hansen2016cmatut}, where the weights $w_i$ decrease logarithmically with the rank of the solution. \modcma provides two alternative options. The first is to use equal weights, $w_i = 1/\mu$ for $i = 1, 2, \ldots, \mu$, effectively performing intermediate recombination without weighting. The second option introduces exponentially decreasing weights, using $w_i = 1/2^{i} + 1/(\mu 2^\mu)$ for $i = 1, 2, \ldots, \mu$. When active update (see Section \ref{modcma:sec:active}) is enabled, we also require weights for the $\lambda - \mu$ individuals that have not been selected. In general, these weights are generated by the same procedure as the positive weights, sequentially reversed and with their sign inverted, i.e. $\mathbf{w}_{-} = (-w_{\mu}, -w_{\mu - 1}, \dots, -w_1)$. For example, for equal weights, assuming $\mu = \lambda / 2$, the complete weight vector equals $[\mathbf{w}; -\mathbf{w}]$.\footnote{For the default weights calculation, this is slightly more involved. See: \citet{hansen2016cmatut}}

\subsection{Sequential Selection}
In standard CMA-ES, each generation samples and evaluates a fixed number $\lambda$ of candidate solutions. 
Sequential selection \citep{brockhoff_mirrored_2010} modifies this process by terminating the generation loop (lines 7--10 in Algorithm \ref{alg:modcma}) early, as soon at least $\mu$ solutions have been evaluated and an improvement over the previous generation has been observed. This approach requires evaluating candidate solutions sequentially, rather than in parallel, but can reduce the total number of function evaluations per generation by avoiding unnecessary evaluations once progress is detected.

\section{Adaptation}
\label{modcma:sec:adaptation}
In this subsection, the components that perform adaptation are discussed in more detail. Specifically, step-size adaptation (SSA) and covariance matrix adaptation (CMA). The latter comprises two modules, active update and matrix adaptation, and is somewhat of a misnomer, since the latter includes modules that do not require an explicit covariance matrix.

\subsection{Step-Size Adaptation}
Supplementary to the default Cumulative Step size Adaptation (\texttt{CSA}), which is shown in lines 13--14 of Algorithm \ref{alg:modcma}, \modcma includes eight alternative mechanisms for performing SSA. These methods can offer advantages over CSA, which struggles with large population sizes $\lambda$ and ineffective variable coordinates \citep{beyer2003qualms}. In the latter case, or when $f(\x)$ is noisy, CSA may lead to too-small step-sizes and premature convergence. 
The following alternatives are implemented:\footnote{We introduce each option on a high level. For implementation details, please refer to the cited papers.}

\begin{enumerate}
    \item Two Point step size Adaption (\texttt{TPA}) \citep{hansen_cma-es_2008}, which resembles \emph{self-adaptation} on a population level, was implemented as an alternative to CSA. The method is less sensitive to algorithm parameters and the properties of $f$, and does not rely on a predefined optimality criterion, such as the 1/5-th success rule \citep{Rechenberg73}. It has the disadvantage of requiring two additional function evaluations, and the step-size adaptation is based solely on these two evaluations, thereby disregarding information from the population. 
    
    \item Success Rule (\texttt{SR}) \citep{Igel2006computational}: SR step-size adaptation uses the success ratio, which
    is an exponential smoothed variable, computed via: $$s = (1 - c_s)s + \left(c_s \mathbf{1}_{f(\x_{1:\lambda}\ng) < f(\x_{1:\lambda}\g)}\right).$$ Here, $f(\x_{1:\lambda}\g)$ denotes the fitness of the best individual in generation $g$. In other words, when the best individual in generation $g+1$ improves over the best individual in generation $g$, the value $s$ increases; it decays with a rate of $1 - c_s$. Based on the classic 1/5-th success rule, $\sigma$ is adapted to match a target success rate of $2/11$~\citep{Igel2006computational}.

    \item Median success rule (\texttt{MSR}) \citep{msr}: Similar to SR-based step-size adaptation, the MSR mechanism uses a notion of success in order to adapt $\sigma$. Specifically, it checks whether the \emph{median} solution candidate, based on $f(\x)$, is improving in successive generations. This is also exponentially smoothed, with a target success rate of 1/2. That is, when $s$ is larger than $1/2$, $\sigma$ is increased, and decreased otherwise. 

    \item Population success rule (\texttt{PSR}) \citep{loshchilov2014computationally}: While both MSR and SR rely on a single value to estimate success over generations, PSR incorporates the function values of the entire population to adapt $\sigma$. It uses a rank-based method: all individuals from the current and previous populations are jointly ranked, then the ranks of each group are extracted. The average rank difference between current and previous populations defines the population success rate, which governs step-size updates.

    \item xNES step-size adaptation (\texttt{xNES}) \citep{glasmachers2010exponential, wierstra2008natural, qualitative_ssa_krause_glas_2017}: The exponential Natural Evolution Strategy (xNES) adapts the step size by analyzing the lengths of normalized mutation vectors, aggregated over the entire population, i.e. $\langle\z\rangle = \sum w_i \z_{i:\lambda}$, to the expected length of a standard normal distribution $\approx \sqrt{d}$. If, on average, the $||\langle\z\rangle||^2$ is longer than expected, the step size is increased; if shorter, it is decreased. 

    \item Mean-xNES step-size adaptation (\texttt{m-XNES}) \citep{qualitative_ssa_krause_glas_2017}: This variant modifies the xNES approach by basing the step-size update on the movement of the search distribution's mean $\m$, rather than aggregating over all offspring. Specifically, the difference between $\m\g$ and $\m\ng$ scaled by $\mu_{\text{eff}}$ is used to update $\sigma$.

    \item xNES with log normal prior Step size adaptation (\texttt{lp-xNES}) \citep{qualitative_ssa_krause_glas_2017}: This method follows the principle of \emph{self-adaptation} by explicitly sampling step sizes for each individual in the population. In each generation, a set of $\lambda$ trial step sizes, $\pmb{\sigma} \in \mathbb{R}^\lambda$, is drawn from a log-normal distribution centered on the current step size $\sigma\g$. Each sampled step size is then used to generate one candidate solution. After selection, the step sizes associated with the top-performing candidates are combined using a log-weighted average $\kappa = \mathbf{w} \cdot \log \pmb{\sigma}$. The step-size is then updated via $\sigma = \sigma^{(1 - c_s)}e^{c_s \kappa}$.
     
    \item Self-Adaptation (\texttt{SA}) \citep{beyer2008covariance}: Another self-adaptive operator, this is a weighted version of the operator for the Covariance Matrix Self-Adaptation (CMSA) ES. Similar to the previous method, lp-xNES, the method samples $\lambda$ individual step sizes $\sigma$ for each candidate solution from a log-normal distribution centered on the current global step size $\sigma\g$. Based on the method from \citet{rscmsa} to correct for elitism, the global step size is updated via:
    $$
        \sigma = \sigma \frac{\prod_{j=1}^{\mu} \sigma_j^{w_i}}{\prod_{j=1}^{\lambda}\sigma_j^{1 / \lambda}}
    $$
    This method uses a lower $\mu = \lfloor \lambda / 4 \rfloor$ by default. 
\end{enumerate}

Note that although several of these methods were explicitly developed for the CMA-ES, a few were not.
Specifically, the SA was developed for the CMSA algorithm~\cite{beyer2008covariance}, and the three xNES methods were originally developed for natural evolution strategies. However, as demonstrated in \citep{qualitative_ssa_krause_glas_2017}, they are not incompatible with the CMA-ES. Nevertheless, these mechanisms may overlap in their behavior, potentially introducing unnecessary complexity from an algorithmic configuration perspective \citep{denobel2021tuning}.

\subsection{Matrix Adaptation}
\label{modcma:sec:matrix}
As explained in detail in Section \ref{pre:sec:cma}, the main strength of CMA-ES lies in its invariance to affine transformations of the search space, achieved by learning the dependencies between the $d$ variables. 
This is done by estimating a positive-definite covariance matrix $\C$, which determines the orientation and scaling of the iso-density contours of the mutation distribution to form an arbitrarily rotated and scaled hyper-ellipsoid.

In the canonical CMA-ES, covariance matrix adaptation is tightly coupled with step-size adaptation via the cumulative step-size adaptation (CSA) mechanism. Specifically, the update of the step-size evolution path (line~13 in Algorithm~\ref{alg:modcma}) requires applying the inverse square root of the covariance matrix, $\C^{-1/2}$. Computing this quantity involves a spectral decomposition of $\C$ into its eigenvectors $\B$ and eigenvalues $\D$ (lines~18 and~20 in Algorithm~\ref{alg:modcma}). Although efficient solvers exist for symmetric positive-definite matrices, this operation has a computational complexity of $\mathcal{O}(d^3)$, which becomes prohibitive as the dimensionality $d$ increases. 

To mitigate this cost, the eigenvalue decomposition is typically performed only intermittently, once every 
$\big(10 d (c_1 + c_\mu)\big)^{-1}$
generations in the standard implementation. Additionally, in \modcma, when \texttt{CSA} is not used as the step-size adaptation mechanism and $\C^{-1/2}$ is therefore not required, $\A$ is computed via a Cholesky decomposition $\A\A^\top = \C$. While this decomposition still scales cubically with $d$, it has a significantly smaller constant factor and is therefore computationally more efficient in practice.

The computational cost of covariance matrix adaptation has motivated the development of several alternative update mechanisms that reduce or avoid expensive matrix decompositions altogether~\citep{ros2008simple,suttorp2009efficient,Igel2006computational,akimoto2016projection}. In \modcma, we implement six alternatives to the standard covariance-based update, each corresponding to a distinct matrix adaptation strategy:

\begin{enumerate}
    \item The first and simplest method is to disable the CMA mechanism entirely. This reduces the algorithm to a variant of CSA-ES \citep{chotard2012cumulative}, in which only step-size adaptation (SSA) is performed and $\C = \mathbf{I}$ remains fixed. When $f(x)$ is separable and well-conditioned, this strategy remains effective.

    \item The separable CMA-ES \citep{ros2008simple} updates only the diagonal matrix $\D = \sqrt{\diag{\C}}$, keeping $\B = \mathbf{I}$ fixed. This reduces the algorithm to linear time complexity while still allowing independent sampling along each coordinate. It can outperform pure step-size adaptation on ill-conditioned problems, and may exceed CMA-ES performance on strictly separable functions. This is not only more computationally efficient than the full CMA update, but also reduces the degrees of freedom in the strategy parameters from $\frac{d^2 + d}{2}$ to $d$. Especially for large $d$, the former can make the CMA-ES highly sample \emph{inefficient}. For example, in direct policy search \citep{wierstra2014natural}, the amount of weights, i.e. $d$, already scales exponentially to the number of nodes in the policy network. 
    In this case, if the adaptation of $\C$ would also require at least $d^2$ samples, the number of samples required for adaptation renders the strategy impractical, even if the decomposition of the covariance matrix might still be technically feasible. 

    \item The Matrix Adaptation Evolution Strategy (MA-ES) \citep{beyer_simplify_2017} is a simplified CMA-ES variant that eliminates both the covariance matrix $\C$ and the evolution path $\pc$ used for the rank-one update. Instead of explicitly maintaining $\C$, MA-ES directly learns a transformation matrix $\mathbf{M}$ such that $\C = \mathbf{M}\mathbf{M}^\top$, using only matrix-vector and matrix-matrix operations. This avoids the expensive spectral or Cholesky decomposition \citep{Igel2006computational}, while empirically retaining performance comparable to standard CMA-ES \citep{beyer_simplify_2017}. Its structural similarity to CMA-ES enables seamless integration with other modules in \modcma, where $\A = \mathbf{M}$.
    
    \item The Cholesky CMA-ES \citep{suttorp2009efficient,Krause2016cma}, avoids perform the expensive spectral decomposition of \C by operating directly on the Cholesky factors $\A\A^T = \C$. In \citet{suttorp2009efficient}, the requirement on $\C^{-\frac{1}{2}}$ was loosened by relying on an approximate conjugate evolution path $\ps$ updated with $\langle \z \rangle_w$. In \modcma, the Cholesky CMA-ES as given in \citet{Krause2016cma} is implemented, which forces the Cholesky factors to be triangular. This ensures a unique and stable decomposition which converges smoothly and is proportional to $\C$. Moreover, the triangular Cholesky factor $\A$ can be inverted in quadratic time, which is also more numerically stable than independently updating $\A^{-1}$ than the approach from \citet{suttorp2009efficient}. Additionally, \citet{Krause2016cma} provided theoretical arguments that using $\A^{-1}$ to update \ps instead of $\C^{-\frac{1}{2}}$ does not impair performance\footnote{Here the matrix \A denotes the triangular Cholesky factor, not $\B\D$ as in Algorithm \ref{alg:modcma}.}.    
    
    \item The Covariance Matrix Self-Adaptation Evolution Strategy (CMSA-ES) \citep{beyer2008covariance} proposes a simplified CMA mechanism that eliminates the evolution paths. This only relies on the points sampled in the current generation; this makes the estimator for $\C$ less stable, but only requires the \emph{rank-$\mu$} update (Line 16 of Alg. \ref{alg:modcma}). The required decomposition of $\C$ is then performed via Cholesky decomposition, yielding a speedup relative to the standard CMA-ES. Notably, the algorithm requires an increased selective pressure, with $\mu = \lfloor \lambda / 4 \rfloor$ by default. The CMSA update can be more robust to noise and performs better with larger population sizes \citep{beyer2008covariance}.
    
    \item Natural Evolution Strategies (NES) \citep{glasmachers2010exponential,wierstra2014natural} are a variation on the standard evolution strategy theme, which uses the natural gradient to update the strategy parameters of a continuous search distribution. The method is analogous to natural gradient ascent, and uses the Fisher information matrix $\mathbf{F}$ estimated from the sampled points to perform an iterative update:
    \begin{equation}
     \theta \gets \theta + \eta\ \mathbf{F}^{-1} \nabla_\theta J 
    \end{equation}
    Here $\theta$ denotes the parameters of the search distribution and $\nabla_\theta J$ the search gradient w.r.t. to $\xspace\theta$. As noticed by \citet{akimoto2010bidirectional}, the so-called exponential NES (xNES) for multivariate normal distributions is closely related to the CMA-ES. In fact, the xNES can be considered a principled alternative of the CMA-ES without evolution paths \citep{glasmachers2010natural}. In \modcma, the matrix adaptation mechanism from xNES is implemented as a module, which updates a transformation matrix:
    \begin{equation}\label{modcma:eq:nes}
        \A \gets \A \cdot \exp(\eta \cdot \nabla_\A J)
    \end{equation}
    Where $\eta$ is a learning rate, and the gradient matrix $\nabla_\A J$ is computed via:
    \begin{align}
        \nabla_\A J &\gets \nabla_{\mathbf{M}} J - \tr(\nabla_{\mathbf{M}} J) / d \mathbf{I}\\  \nabla_{\mathbf{M}} J &\gets \sum_{i=1}^{\mu} w_i (\z_i \z_i^\top - \mathbf{I})
    \end{align}
    From a computational perspective, xNES does not offer any benefit over the standard CMA-ES, as the matrix exponential in Equation \ref{modcma:eq:nes} incurs a similar cubic time complexity. 
\end{enumerate}

\paragraph{CSA without $\C^{-\frac{1}{2}}$}
Several matrix adaptation methods presented above do not explicitly perform an eigendecomposition and therefore do not provide an efficient way to compute $\C^{-\frac{1}{2}}$, which is required in the standard CSA update. However, the mutation vectors $\z_i$ are already available from the sampling step: each candidate solution is generated by first sampling $\z_i$ and then applying the transformation $\y_i = \A \z_i$. Hence, instead of whitening the selected mean step $\langle \y \rangle_w$ by multiplying it with $\C^{-\frac{1}{2}}$, the CSA update can use the weighted average of the corresponding normalized mutation vectors directly,
\begin{equation}
    \langle \z \rangle_w = \sum_{i=1}^{\mu} w_i \z_{i:\lambda},
\end{equation}
where the $\z_i$ are sorted according to $f(\x_i)$. The step-size evolution path is then updated as
\begin{equation}
    \ps \gets (1 - c_\sigma)\ps
    + \sqrt{c_\sigma(2-c_\sigma)\mu_{\text{eff}}}\,\langle \z \rangle_w .
\end{equation}

For the \texttt{None} and \texttt{Separable} methods, this is exact because the sampling transformation does not introduce a rotated coordinate system. For methods that maintain a transformation matrix $\A$, such as Cholesky CMA-ES and MA-ES, the relation $\langle \z \rangle_w = \A^{-1}\langle \y \rangle_w$ holds by construction \citep{beyer_simplify_2017,Krause2016cma}. Since the vectors $\z_i$ are already known, this inverse transformation does not need to be computed explicitly. This avoids an additional matrix-vector product and the computation of $\A^{-1}$.

\subsection{Active Update}
\label{modcma:sec:active}
Active-CMA-ES \citep{jastrebski2006improving} extends the standard CMA-ES by incorporating information from both successful and unsuccessful candidate solutions into the covariance matrix adaptation process. While traditional CMA-ES increases variances in directions associated with successful steps and allows other directions to decay passively, Active-CMA-ES actively reduces variances in directions associated with poor solutions. This is accomplished by modifying the rank-$\mu$ update term to include negative contributions from the worst-performing individuals. Specifically, the update in line 16 from Algorithm~\ref{alg:modcma} 
\begin{equation}
    \C \gets [\dots] +  c_\mu \sum_{i=0}^{\color{orange}\mu} w_i \y_{i:\lambda} (\y_{i:\lambda})^\top
\end{equation} 
is extended to
\begin{equation}
     \C \gets [\dots] +  c_\mu \sum_{i=0}^{\color{orange}{\lambda}} w_i \y_{i:\lambda} (\y_{i:\lambda})^\top
\end{equation}
where the weights $w_i$ assigned to the $\lambda - \mu$ worst individuals are negative. This active update accelerates the adaptation of the covariance matrix and has been shown to empirically improve performance on standard benchmark problems \citep{jastrebski2006improving}. Note that in \citet{hansen2016cmatut}, this variant is treated as the default; here, we consider it an optional modification. For canonical xNES, active update is the standard, since the update of $\nabla_{\mathbf{M}}$ is originally defined as a sum over all $\lambda$ individuals \citep{wierstra2014natural}. Here, the default is to sum over $\mu$, and with active update enabled, to sum over $\lambda$.
 
\section{Restarting}
\label{modcma:sec:restarting}
When the optimization process stagnates, the CMA-ES can be restarted using a restart strategy \citep{auger2004ls}. By default, the restart criteria specified in \citet{hansen2016cmatut} are used. These include checks on the internal parameters, such as $\sigma$ and $\C$, and stagnation of the observed $f(\x)$ values. In \modcma, any of these criteria can be disabled or their tolerances modified to suit the user's needs. Moreover, custom restart criteria can be specified, and potentially combined, providing greater flexibility. If any of these criteria evaluates to true, the CMA-ES will be stopped or restarted. What happens following a restart is controlled by the next two modules.

\subsection{Restart Strategy}
The restart strategy specifies how the algorithm proceeds once a restart condition is triggered. The simplest approach is to ignore the restart signal and continue the optimization. While this may lead to numerical issues if the algorithm is run indefinitely, it can be useful for standardized experiments with a fixed evaluation budget. The second approach, \texttt{Restart}, resets the CMA-ES using the same parameters as in the initial run \citep{auger2004ls}. The initial position $\mathbf{x}_0$ can be reused or resampled uniformly from $[\lb, \ub]^d$. This flexibility is also present in the next two strategies:

\begin{enumerate}
    \item Increasing Population (\texttt{IPOP}) \citep{auger2005restart}: In order to achieve fast convergence on simple problems, the CMA-ES has a relatively small population size compared to traditional ES \citep{Beyer01,hansen2001completely}. However, on multimodal functions, a larger population size $\lambda$ is required to solve them effectively \citep{Hansen2004Evaluating,Lunacek2008impact}. The IPOP strategy first attempts to optimize with the default small $\lambda$, then increases $\lambda$ by a factor of 2 at each restart. The population size is limited to a maximum of $2^9\lambda_0$, where $\lambda_0$ denotes the original population size specified by the user. This maintains competitive performance on simple functions that do not require restarts, while improving performance on complex multimodal landscapes in subsequent restarts. 

    \item Bi-Population (\texttt{BIPOP})~\citep{bipop}: BIPOP extends the IPOP concept by alternating between two regimes: one with an increasing population size (as in IPOP), and one with a smaller, randomly parameterized population size. Each regime tracks its own evaluation budget independently. On each restart, the regime with fewer total evaluations is selected. Initially, the CMA-ES is started with standard parameters, and the increasing regime is used for the first restart. The population size for the small regime is sampled on each restart as:
    \begin{equation}
        \lambda = \left\lfloor \lambda_0 \left( \frac{1}{2} \cdot \frac{\lambda_l}{\lambda_0} \right)^{\mathcal{U}(0, 1)^2} \right\rfloor
    \end{equation}
    where $\lambda_l$ is the current population size of the large regime, and $\mathcal{U}(0, 1)$ is a uniform random variable. Additionally, the step size in the small regime is randomized as $\sigma = \sigma_0 \times 10^{-2 \cdot \mathcal{U}(0, 1)}$. The balance between the exploratory, increasing, and exploitative smaller regimes yields a highly effective strategy. The BIPOP CMA-ES is currently one of the top-performing algorithms on BBOB \citep{hansen2022anytime}.
\end{enumerate}
 
\subsection{Repelling}
Native restart strategies can be effective for multimodal landscapes but typically ignore information gained from previous runs. In contrast, state-of-the-art evolutionary algorithms for multi-modal optimization (MMO), such as HillValley-EA \citep{maree2018real} and RS-CMSA \citep{rscmsa}, explicitly leverage such historical knowledge. Inspired by these approaches, the Repelling Restarts CMA-ES \citep{denobel2024avoiding} maintains a taboo archive of previously visited basins of attraction and actively repels new candidate solutions away from these regions. This mechanism is updated on each restart and can be combined with any of the restart schemes described above.

\section{Auxiliary Components}\label{sec:mod:other}
The following subsections describe components of the \modcma that are not strictly encoded as modules but work in conjunction with the framework's modular structure.   

\subsection{(Mixed)-Integer Optimization}
The \modcma incorporates a strategy for handling integer variables that is inspired by the lower-bounding approach proposed by \citet{marty2024lb}. In the original work, two mechanisms are introduced to improve the treatment of integer variables within CMA-ES; however, in our implementation, we restrict ourselves to the lower-bounding component only. Concretely, for each integer variable, the effective step size $v_i = \sigma \cdot\C_{(i,i)}$ is prevented from becoming too small by imposing the lower bound $\min\left(\frac{1}{5}, \frac{\mu_{\text{eff}}}{d}\right)$. This ensures that sufficient variation is maintained along integer dimensions and avoids the coordinate variances shrinking to far below the integer bound. Additionally, during the sampling step, each integer variable is rounded to its nearest integer value, i.e., immediately after line 10 in Algorithm~\ref{alg:modcma}, effectively generating integer solutions according to a truncated (normal) distribution \citep{shir2025}.

\subsection{The (1+1)-CMA-ES}

Although not represented by a separate module, setting the population size to $\lambda=1$ activates the $(1+1)$-CMA-ES mode, based on the procedure described in~\citet{Igel2006computational}. In this setting, the single offspring is compared directly against its parent, and the distribution parameters are updated only when the offspring is successful.

Several module choices are necessarily fixed or disabled in this regime. Specifically, $\mu$ is fixed to one, elitist selection is enabled, \emph{Sequential Selection} and \emph{Active Update} are disabled, and the recombination weights are fixed to equal weighting. The step-size adaptation mechanism is also fixed to the success-rule-based adaptation used by the $(1+1)$-CMA-ES. Finally, population-based restart schemes such as \texttt{IPOP} and \texttt{BIPOP} reduce to a standard restart, since increasing or alternating population sizes is not applicable to the $(1+1)$ setting. These restrictions ensure that configurations that have no meaningful interpretation for a single-offspring strategy do not result in separate algorithm variants.

\subsection{\texttt{ConfigSpace}}
\label{sec:configspace}

To expose the many modules, options, and parameters of \modcma in a way that is compatible with modern hyperparameter optimization (HPO) frameworks, the package provides a \texttt{ConfigSpace} interface~\citep{configspace}. The resulting configuration space contains one categorical hyperparameter for each module listed in Table~\ref{modcma:tab:options}, where the possible values correspond to the available module options. In addition, the interface can expose a selected set of numerical parameters controlling population size, initial step size, and the main learning rates of the CMA-ES update. This structured design space allows users to conveniently tune both high-level algorithmic choices and lower-level numerical settings, while ensuring compatibility with HPO tools such as \textsc{SMAC}. Table~\ref{tab:configspace_numeric} lists the numerical parameters currently included in the \texttt{ConfigSpace} interface. The population-size parameters are represented as log-scaled integer hyperparameters, with the additional constraint $\mu_0 \leq \lambda_0$. The remaining numerical parameters are represented as bounded continuous hyperparameters centered on their default values. Unless stated otherwise, the defaults are computed from the standard \modcma parameter initialization for the specified dimension, see~\citet{hansen2016cmatut} for more details.

\begin{table}[!htbp]
\centering
\small
\caption{Numerical parameters exposed by the \texttt{ConfigSpace} interface of \modcma under the default module configuration. Here $d$ is the search-space dimension, $\lambda_0$ is the initial offspring population size, $\mu_0$ is the number of selected parents,
and $\mu_{\mathrm{eff}} = (\sum_i^{\mu}{w_i^2})^{-1}$ is computed from the normalized positive recombination weights. The bounds for $\lambda_0$ and $\sigma_0$ are practical numerical limits rather than theoretical constraints; extreme values, particularly for $\sigma_0$, are generally better handled through appropriate scaling or normalization of the search space.
}
\label{tab:configspace_numeric}
\begin{tabularx}{\linewidth}{@{} l l l l l X @{}}
\toprule
\textbf{Name} & \textbf{Type} & \textbf{LB} & \textbf{Default} & \textbf{UB} & \textbf{Description} \\
\midrule
$\lambda_0$ 
& $\mathbb{Z}$ 
& $1$ 
& $4 + \lfloor 3 \log d \rfloor$ 
& $50d$ 
& Initial offspring population size. \\

$\mu_0$ 
& $\mathbb{Z}$
& $1$ 
& $\lfloor \lambda_0 / 2 \rfloor$ 
& $\lambda_0$ 
& Initial number of selected parents. \\

$\sigma_0$ 
& $\mathbb{R}$ 
& $10^{-15}$ 
& $2.0$ 
& $10^{15}$ 
& Initial global step size. \\

$c_\sigma$ 
& $\mathbb{R}$ 
& $0$ 
& $\dfrac{\mu_{\mathrm{eff}} + 2}{d + \mu_{\mathrm{eff}} + 3}$ 
& $1$ 
& Learning rate of the step-size evolution path. \\

$c_c$ 
& $\mathbb{R}$ 
& $0$ 
& $\dfrac{4 + \mu_{\mathrm{eff}}/d}{d + 4 + 2\mu_{\mathrm{eff}}/d}$ 
& $1$ 
& Learning rate of the covariance evolution path. \\

$c_\mu$ 
& $\mathbb{R}$ 
& $0$ 
& $\min\left(1-c_1,\,
\dfrac{2\left(\mu_{\mathrm{eff}} - 2 + 1/\mu_{\mathrm{eff}}\right)}
{(d+2)^2 + \mu_{\mathrm{eff}}}\right)$ 
& $1$ 
& Learning rate of the rank-$\mu$ covariance update. \\

$c_1$ 
& $\mathbb{R}$ 
& $0$ 
& $\dfrac{2}{(d+1.3)^2 + \mu_{\mathrm{eff}}}$ 
& $1$ 
& Learning rate of the rank-one covariance update. \\

$d_\sigma$ 
& $\mathbb{R}$ 
& $0$ 
& $1 + 2\max\left(0,\sqrt{\dfrac{\mu_{\mathrm{eff}}-1}{d+1}} - 1\right) + c_\sigma$ 
& $10$ 
& Damping factor for cumulative step-size adaptation. \\
\bottomrule
\end{tabularx}
\end{table}

\section{A closer look at Matrix Adaptation}
\label{modcma:sec:matrix:exp}
Matrix adaptation is arguably the defining module of the \modcma framework. Viewed as a variable-metric method, CMA-ES derives much of its behavior from the metric learning during the search. Moving away from the standard covariance matrix adaptation mechanism alters the geometry of the search and, with it, the character of the strategy itself.

In this section, we examine the options in greater detail through two experiments. First, we consider the computational perspective, measuring the duration of the adaptation step. Secondly, we compare the different options on a set of benchmark problems 
As a baseline, we use the official CMA-ES implementation from the \emph{Pycma} Python library (version 4.2.0)  
in both experiments.\footnote{\emph{Pycma} is available at \url{https://github.com/CMA-ES/pycma}.} In order to get comparable results, \emph{active update} is disabled in all algorithms, and all parameters, such as $\mu$ and $\lambda$, are kept consistent betwen the compared packages. For \modcma (version 1.1.0),
only the matrix adaptation mechanism is modified; the other options are set to their default values (see Table \ref{modcma:tab:options}).

\subsection{Running Time}
\label{modcma:sec:matrix:runtime}
We measure the median (single-threaded) running time for each algorithm to complete 1\,000 generations. The objective function used is a $d$-dimensional ellipsoid, i.e., BBOB $f_2$. For every $d\ \in \{2, 3, 5, 10, 20, 50, 100, 200, 500, 1\,000\}$, 15 runs are conducted. The runtime was measured using Python 3.10.12 on a desktop machine running the Windows subsystem for Linux (version 6.6.87.1) with a 12th Gen Intel i5-12400F CPU.

\begin{figure}[!htbp]
    \centering
    \includegraphics[width=\textwidth]{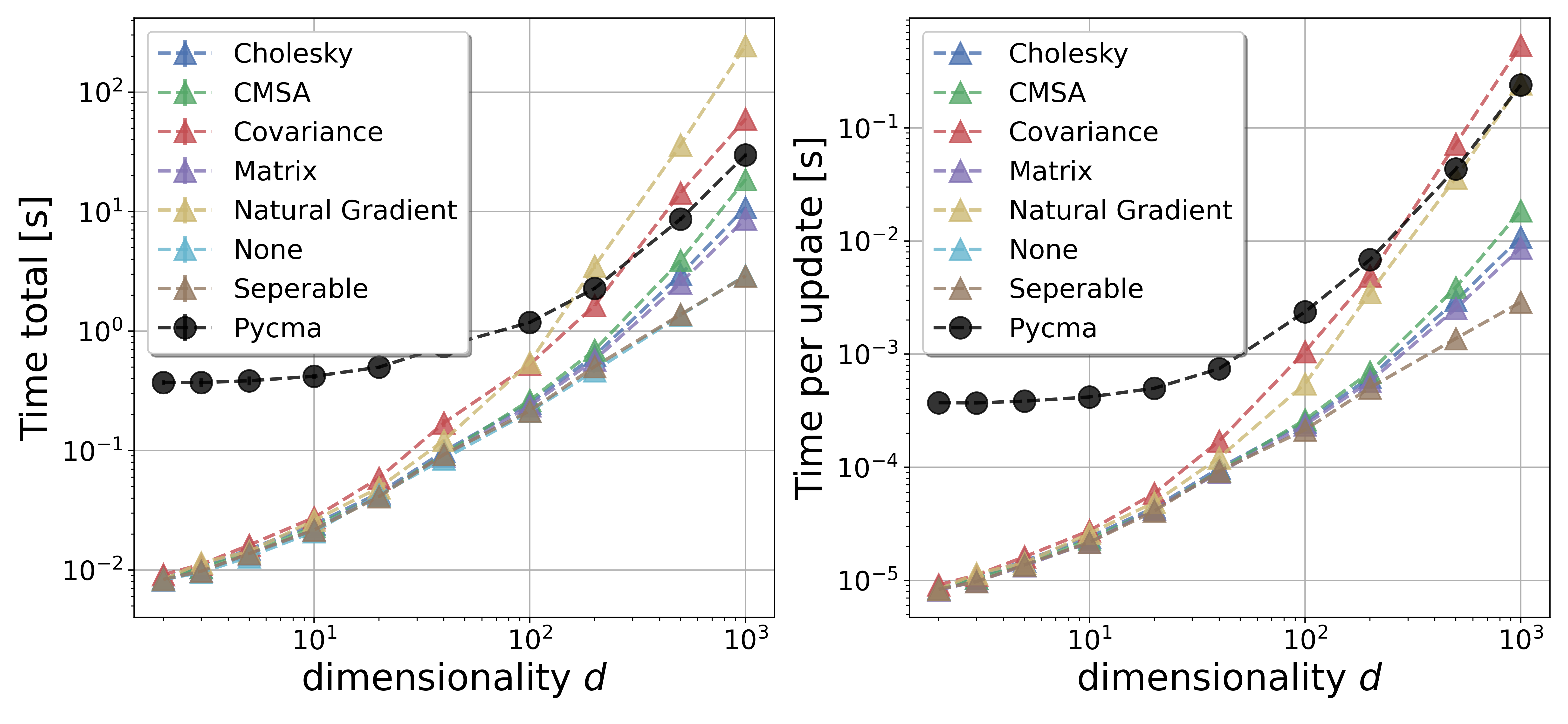}
    \caption{Running time in seconds vs. search space dimensionality $d$. The left figure shows the total time to perform 1\,000 generations of the algorithm, using the default population size of $4 + \lfloor 3 \log d \rfloor$. The median time over 15 independent runs is shown. The right figure shows the average time per matrix update. }
        \label{fig:runtime}
\end{figure}

Figure~\ref{fig:runtime} (left) shows the median time in seconds for each algorithm to complete the required number of generations. The right figure shows the normalized time per update. This is added because the standard CMA (\texttt{Covariance}) does not update its strategy parameters at each generation. This is evident when comparing the relative performance of \texttt{Covariance} with the other options in the right and left figures. Especially for higher values of $d$, the time-per-update difference is much larger. Overall, the most computationally demanding matrix adaptation method is the \texttt{Natural Gradient} option based on xNES. It should again be noted that this holds only for the total time. When considering the time per update, the standard CMA method is the least efficient among the options. The scaling of the time per update is similar to the \texttt{Natural Gradient} method; the constant for performing a full spectral decomposition of \C seems higher than that for computing a matrix exponential. If we compare this with the \texttt{CMSA} method, which uses a Cholesky factorization rather than an eigen decomposition, we already observe a significant speedup. In fact, \texttt{CMSA} with a Cholesky decomposition is only marginally slower than the Cholesky CMA-ES, which doesn't perform a decomposition. The \texttt{Matrix} adaptation method is slightly faster than the Cholesky CMA-ES. The two fastest methods, \texttt{None} and \texttt{Separable}, are indistinguishable from each other in Figure~\ref{fig:runtime}. If we compare these results against the baseline CMA-ES implementation \emph{Pycma}, we can observe that all methods in \modcma are significantly faster for $d \leq 100$. However, for larger $d$, once the eigenvalue decomposition starts to dominate the computation, the runtime becomes similar to that of the \texttt{Covariance} matrix adaptation method implemented in \modcma. In fact, for $d > 500$, the implementation in \emph{Pycma} is notably faster, both overall and per update. This might because the eigensolver of \textsc{NumPy} is more optimized for large matrices than then one included in \textsc{Eigen}.\footnote{\textsc{np.linalg.eigh} is the default eigensolver in \emph{Pycma}. This uses a platform-specific numeric LAPACK routine. For \modcma, the \textsc{SelfAdjointEigenSolver} from \textsc{Eigen} is used.} Nevertheless, faster implementations, such as the Cholesky CMA, significantly outperform \emph{Pycma}
in runtime across all tested values of $d$.

\subsection{Benchmarking}
\label{modcma:sec:matrix:benchmark}
In this section, benchmark data is collected for the benchmark functions taken from the \textbf{BBOB} suite, implemented by \IOHexp \citep{de2024iohexperimenter}. As in the previous section, only the different matrix adaptation methods are studied, with all other options set to their default values. For each function and each dimensionality $d \in \{2, 3, 5, 10, 20, 40\}$, 100 independent trials are run for every variant. Only the first instance is used, and each algorithm is initialized at the origin with $\sigma_0 = 2.0$. No restarting is allowed, and each run is stopped once the final target value of $10^{-9}$ has been reached, any of the internal restart criteria have been met, or after the maximum budget of $B = 10^4 d$ has been used up by the algorithm.

\paragraph{Expected Running Time (ERT)} \label{metric:ert} 
We report the \emph{expected running time} (ERT). Let $\Delta f = f(x) - f^\star$ denote the precision to the optimum. For a given target precision $\Delta f_{\text{target}}$, ERT measures the expected number of function evaluations required to reach $\Delta f \leq \Delta f_{\text{target}}$. Specifically, we have $R$ independent runs, and let $s$ denote the number of runs that successfully reach the target.
Then, let $T_i$ be the number of function evaluations required to reach the target if successful, and let $B$ denote the evaluation budget for unsuccessful runs. The ERT is then defined as
\begin{equation}
\mathrm{ERT} = \frac{\sum_{i=1}^{R} T_i + \sum_{i=1}^{R-s} B}{s},
\end{equation}
Lower ERT values indicate better performance, and, if $s = 0$, ERT = $\infty$.

\begin{figure}[!htbp]
    \centering
    \includegraphics[width=\textwidth]{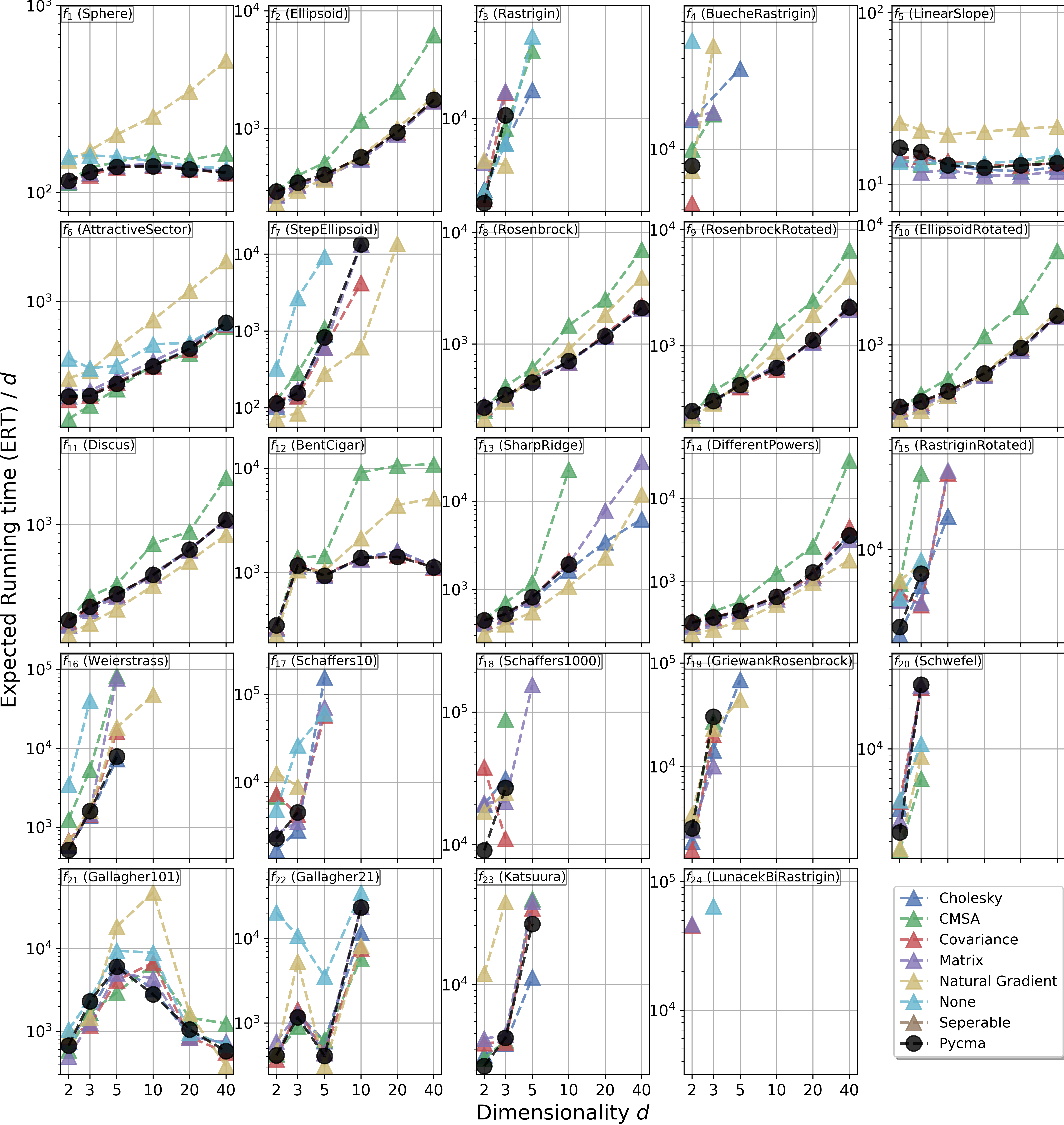}
    \caption{$\log_{10}$ Expected Running Time (ERT) divided by the dimensionality $d$ for the benchmark functions from the BBOB suite, for six different values of $d$.  All matrix adaptation methods are compared with the official Python CMA-ES implementation \emph{Pycma}. The ERT is computed over 100 independent runs, each with a maximum budget of $10^4 d$, to achieve a target value of $10^{-9}$. Lower values are better.} 
        \label{fig:benchmark}
\end{figure}

\paragraph{Results} 
Figure~\ref{fig:benchmark} presents the ERT achieved by each matrix adaptation mechanism on the benchmark functions, for a target precision of $\Delta f_{\text{target}} = 10^{-9}$. The official \textsc{cma} implementation (\emph{Pycma}) is included as a baseline. Ideally, we expect its performance to align with that of the \texttt{Covariance} strategy in \modcma, as both implement the same algorithm. However, \emph{Pycma} is a highly optimized and customized version of CMA-ES. It includes several implementation-specific enhancements, such as mechanisms for handling neutrality and repairing strategy parameters, none of which are present in \modcma.

Despite these differences, the empirical performance of \modcma with \texttt{Covariance} matrix adaptation aligns well with \emph{Pycma} across most benchmarks. Minor differences can be observed on multimodal functions, such as $f_4$, $f_{15}$, $f_{18}$, and $f_{21}$. However, this variability is likely due to the inherent randomness in optimizing multimodal functions with sample-based optimization heuristics. 

Turning to the other matrix adaptation strategies, several patterns emerge. On functions $f_1$ and $f_6$, the \texttt{None} strategy (i.e., no matrix adaptation) performs well, and on $f_3$ its performance is comparable to that of the other strategies. However, since all methods struggle with $f_3$, a highly multimodal function, meaningful performance differences are hard to observe and analyze. On other multimodal functions, such as $f_{17}$ and $f_{21}$, the \texttt{None} strategy also seems to perform as well as the other methods, suggesting that on these landscapes, matrix adaptation is less important.  

The \texttt{Separable} strategy, as expected, excels on separable functions but fails on non-separable ones. A clear illustration of this is the performance contrast between $f_2$ (separable) and $f_{10}$ (a rotated, non-separable version of $f_2$). The \texttt{Separable} method performs well on $f_2$ but poorly on $f_{10}$, highlighting its limitations. 

Among the more computationally efficient matrix adaptation methods, \texttt{Cholesky} and \texttt{Matrix} show no significant performance drop across most benchmarks. In fact, both the \texttt{Cholesky} and \texttt{Matrix} adaptation methods are able to find optimizers for $d> 10$ on $f_{13}$, which neither the \texttt{Covariance} nor the \emph{Pycma} methods are able to do. While already clearly demonstrated in the original works \citep{beyer_simplify_2017,Krause2016cma}, each of these methods is indeed a viable alternative to the computationally more demanding \texttt{Covariance} method. 

In general, the \texttt{CMSA} strategy tends to be somewhat slower across the board in terms of ERT. The same holds for the \texttt{Natural Gradient} approach. Importantly, neither of these methods is tailored for the \texttt{CSA} step-size adaptation strategy, which is used for all methods in this evaluation. This mismatch may partially explain their comparatively lower performance. Still, it is worth noting that on functions $f_7$, $f_{11}$, $f_{13}$, and $f_{14}$, the \texttt{Natural Gradient} method often achieves the lowest ERT among all evaluated strategies. 

Overall, without restarts and with the default population size, many of the benchmarks shown in Figure~\ref{fig:benchmark} are not solvable even in moderately high dimensions within our configured budget. For example, on $f_{24}$, no meaningful conclusions can be drawn, since almost no runs across all algorithms reach the final target value of $10^{-9}$. Interestingly, the multimodal function $f_{21}$, however, seems to get more solvable with increasing dimension. These results highlight that this comparison is primarily informative for problems and dimensions where the considered configurations can reliably reach the target within the available budget.

\section{Automated Configuration}
\label{modcma:sec:hyper}

To evaluate the benefits of the modular design, we study how well \modcma can be automatically configured for different optimization problems. Rather than proposing a single ``best'' configuration, we explicitly investigate whether tailoring the algorithm to individual problems yields measurable improvements over the default CMA-ES design. Note that these experiments serve primarily to illustrate the use of \modcma for automated configuration and specialization to individual problems; a comprehensive tuning study is beyond the scope of this paper.

\subsection{Tuning Setup}
\label{modcma:sec:tuning_setup}

As before, we consider the noiseless BBOB test suite~\cite{hansen2020coco}. All experiments in this section are conducted in dimension $d=5$, using the first $10$ instances for the configuration process, with a maximum of $25$ runs for each algorithm configuration across these instances. 

\paragraph{Parameter Space for Tuning.}
The search space used for automated configuration is derived from the set of modules in Table~\ref{modcma:tab:options}. We consider the full configuration space with the following modifications:
\begin{itemize}
    \item \textbf{Boundary handling:} The \emph{Boundary Correction} module is excluded, as the BBOB problems are defined on unbounded domains.     
    \item \textbf{Population size:} The population size parameters $\lambda$ and $\mu$ are included as tunable integer parameters in $[1,50d]$,
    with the constraint $\mu \leq \lambda$. 
\end{itemize}

All remaining modules and parameters correspond directly to those listed in Tables~\ref{modcma:tab:options} and \ref{tab:configspace_numeric}.
The resulting configuration space is constructed programmatically using the \textsc{ConfigSpace} interface provided by \modcma, ensuring that all sampled configurations correspond to valid algorithm instances. 

\paragraph{Algorithm Configuration.} 
We use the \textsc{SMAC} framework~\cite{smac} to perform per-function configuration. For each BBOB function, \textsc{SMAC} is run independently with a budget of $50\,000$ configuration evaluations. Each evaluation corresponds to running a \modcma configuration on a single problem instance. \textsc{SMAC} employs an intensification procedure in which promising configurations are evaluated on multiple instances. In our setup, each configuration is evaluated up to $25$ times on different BBOB instances and random seeds, and selection is based on the aggregated performance across these runs. 

\paragraph{Configuration Objective.}
During the \textsc{SMAC} runs, configurations are evaluated using a normalized log-precision measure derived from the best-so-far function values. For each run, precision values are clipped to the interval $[10^{-8},10^{2}]$, transformed to a logarithmic scale, and averaged over the evaluation budget. This yields a scalar performance metric that reflects the convergence speed of a configuration: the area over the convergence curve, equivalent to the area under the empirical cumulative distribution~\cite{lopez2024using}. This objective is to be minimized, such that lower values correspond to faster convergence.

\paragraph{Selection and Validation.}
After optimization, we take only the configurations that have been evaluated for 25 repetitions by the intensification procedure. The configurations are ranked by their average \textsc{SMAC} objective value, and the five best configurations per function are retained. Each of these candidates is then evaluated on $50$ independent BBOB instances that were not used during tuning. For each function, the reported tuned configuration is the candidate with the lowest ERT on this validation set. 

\subsection{Benchmark Results}
\label{modcma:sec:benchmark_results}

For the performance comparison, we again use ERT (see Section \ref{metric:ert}), with a predefined target precision of $10^{-8}$ following the standard COCO/BBOB~\cite{hansen2020coco} methodology\footnote{While we tune for anytime performance in the AOCC, we still test for fixed-target performance, following the observations from~\cite{ye2022automated} which found that AOCC provides a stronger tuning signal. }.
We compare three settings:
\begin{enumerate}
\item \textbf{Default:} the default \modcma configuration.
\item \textbf{Tuned:} the per-function tuned configurations obtained via \textsc{SMAC}.
\item \textbf{bbob2009:} a reference baseline constructed from the per-function best-performing algorithms in the BBOB-2009 benchmark~\cite{hansen2010comparing}.
\end{enumerate}
For both the `Default' and `Tuned' \modcma configurations, we calculate ERT over 50 validation runs. For the reference algorithm, we extract all data from the BBOB-2009 benchmark from the COCO data archive \cite{coco_bbob_data_archive}, and calculate ERT over 15 runs. For each function, the virtual best-solver, i.e., the solver with the lowest ERT, is selected for comparison.

\begin{figure}[!htbp]
    \centering
    \includegraphics[width=\linewidth]{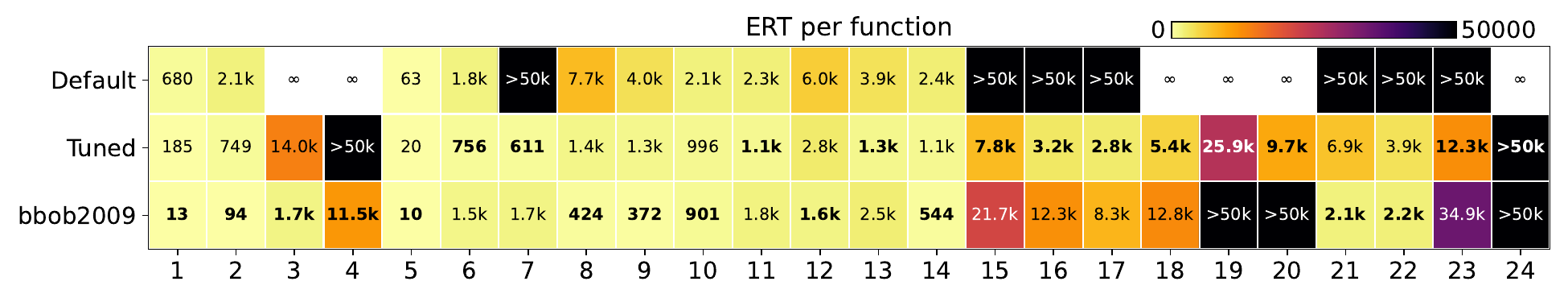}
    \caption{
    Expected running time (ERT) of default and tuned \modcma configurations on the 24 BBOB functions in dimension $5$, compared to the best-performing solver per function from the \textbf{bbob2009} portfolio. Lower values indicate better performance. The best (lowest) value per function is shown in bold. For some functions, configurations fail to reach the target within the evaluation budget, resulting in large or infinite ERT values.}
        \label{fig:ERT_5D}
\end{figure}

\paragraph{Results}
Figure~\ref{fig:ERT_5D} shows the ERT values for all 24 BBOB functions. A larger version of this figure, including individual entries for each baseline algorithm, is provided in Appendix~\ref{sec:ERT_full}. The benefit of automated configuration is clearly visible across the benchmark.

The default configuration fails to consistently reach the target on a substantial subset of functions (e.g., $f_3$, $f_4$, and $f_{15}$–$f_{24}$), resulting in infinite or very large ERT values. In contrast, the tuned configurations substantially reduce ERT on most functions, often by orders of magnitude. For some functions, the improvement is more modest, such as on $f_2$ and $f_9$–$f_{14}$, but even in these cases the tuned configuration improves upon the default by at least a factor of two. On several functions where the default has very low success (e.g., $f_{15}$–$f_{23}$), the tuned configurations reliably reach the target, yielding finite ERT values.

The configurator failed to identify a reliable solver for only two functions. Although the ERT is finite; $120{,}873$ for $f_4$ and $246{,}439$ for $f_{24}$, not all $50$ validation runs reach the target precision of $10^{-8}$, corresponding to success rates of $\frac{38}{50}$ and $\frac{6}{50}$ for $f_4$ and $f_{24}$, respectively. However, these functions are particularly difficult. In the \textbf{bbob2009} reference set, only three out of thirty algorithms achieve a finite ERT on $f_{24}$, all with worse performance than the tuned \modcma. Similarly, for $f_4$, only two algorithms outperform the tuned configuration, namely \textsc{DASA}~\cite{korovsec2009stigmergy} and \textsc{LSstep}~\cite{povsik2009bbob}, both of which are tailored to exploit the separable structure of the problem.

Overall, the tuned \modcma matches or improves upon the \textbf{bbob2009} reference on exactly half of the functions. The gains are most pronounced on complex multimodal problems (e.g., $f_{15}$–$f_{20}$), where performance improves by factors of $2.5$ to $5$. On functions where the reference performs better, the gap is typically small or attributable to highly specialized algorithms. Examples include BFGS on $f_1$ and line-search methods~\cite{povsik2009bbob} on $f_2$ and $f_3$, which further motivates hybrid approaches combining CMA-ES with such techniques~\cite{loshchilov2013bi}.

\subsection{Interpreting the Configuration Space}
\label{modcma:sec:shap}

While the tuning results show that \modcma can adapt effectively to different problem characteristics, they do not directly reveal which components of the algorithm drive these improvements. To gain further insight, we analyze the configurations evaluated during the SMAC \cite{smac} runs using Shapley-based feature attribution \cite{shap}, following the setup of~\cite{kononova2025xai}.

\paragraph{Data collection.}
During the SMAC optimization process, we evaluate many configurations. Each evaluation yields a configuration together with its performance. We collect these for all functions into a dataset of configuration--performance pairs, using the internal SMAC cost (AOCC) as the performance measure.

\paragraph{Surrogate model.}
To enable feature attribution, we train a regression model that predicts a configuration's performance from its parameter settings. We use a \textsc{CatBoost} regressor~\cite{catboost}, a gradient-boosting model designed for handling categorical variables. We train a separate model for each function on the full set of evaluated configurations. 

\paragraph{\textsc{SHAP} value computation.}
Given the trained surrogate model (\textsc{CatBoost}), we compute feature attributions using the \textsc{TreeSHAP} method~\cite{shap}. For each evaluated configuration, \textsc{SHAP} assigns a contribution value to every parameter, indicating how much that parameter influences the performance. Note that a negative contribution value indicates a positive impact on performance, as the cost is minimized. \textsc{SHAP} values are expressed in the same scale as the surrogate model output, i.e., the \textsc{SMAC} cost.

\begin{figure}[!htbp]
\includegraphics[width=\textwidth]{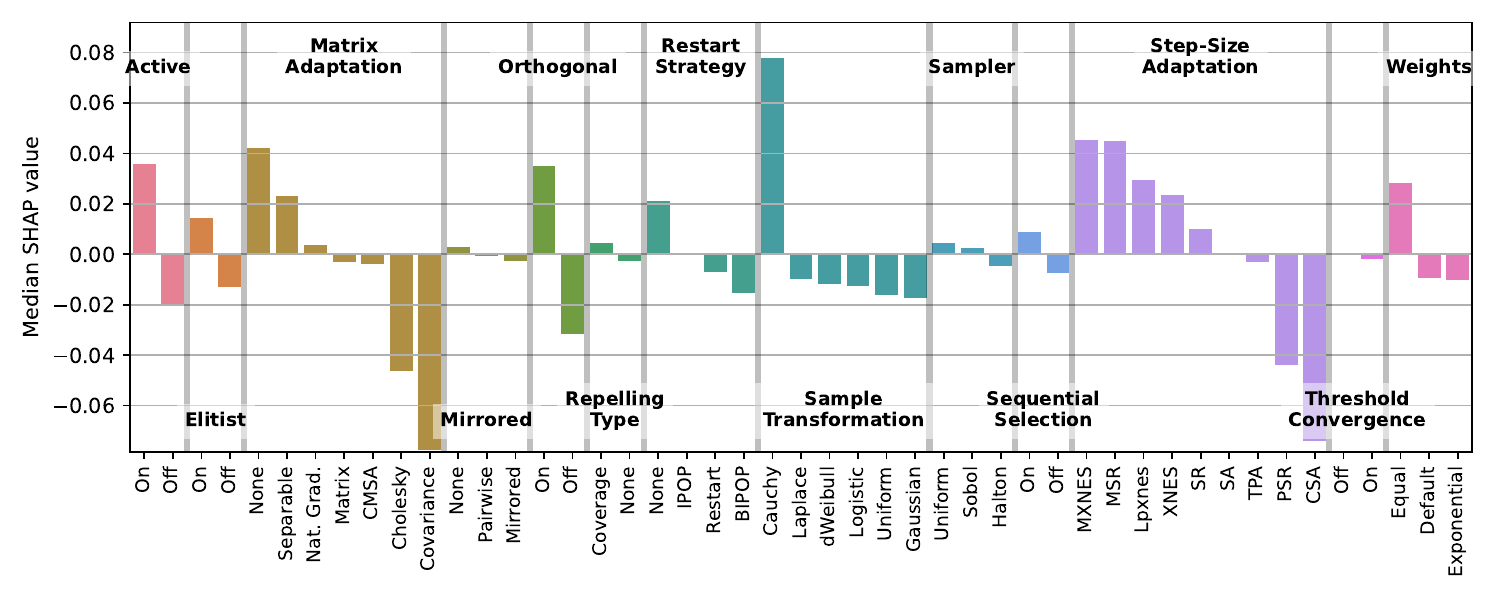}
\caption{Global contribution of module options based on \textsc{SHAP} values. Each bar shows the median SHAP value of a configuration option, aggregated over all evaluated configurations from the \textsc{SMAC} runs. Options are grouped by module (separated by vertical lines), with module names shown as annotations. Negative values indicate options that tend to improve performance, while positive values indicate options that tend to degrade performance, according to the surrogate model.}
\label{fig:shap_global}
\end{figure}

\paragraph{Global importance.}
To assess overall importance, we aggregate \textsc{SHAP} values across all configurations by computing the median value per option, as shown in Figure~\ref{fig:shap_global}. Several modules, such as \emph{Step-Size Adaptation} (SSA), strongly influence performance, whereas others, such as repelling restart, have only a marginal effect. As these results are aggregated over all 24 BBOB functions, options that are highly beneficial for specific functions may appear less influential overall, and mechanisms that are only active under certain conditions (e.g., repelling restarts requiring a restart strategy) naturally show reduced impact~\citep{denobel2024avoiding}. The sign of the SHAP values indicates whether options tend to improve or degrade performance; for example, the binary modules Active and Orthogonal have slightly more positive than negative SHAP values overall, suggesting a modest negative effect on performance. However, these values reflect average effects within the explored configuration space and are influenced by interactions with other parameters, so options that are beneficial in isolation may appear detrimental on average. Finally, we observe substantial variation within modules: for instance, within SSA, \texttt{CSA} generally contributes positively, whereas \texttt{MXNES} tends to be negative, highlighting that both the specific option and the interplay with other modules \cite{denobel2021tuning} play a key role in determining performance.

\begin{figure}[!htbp]
\includegraphics[width=\textwidth]{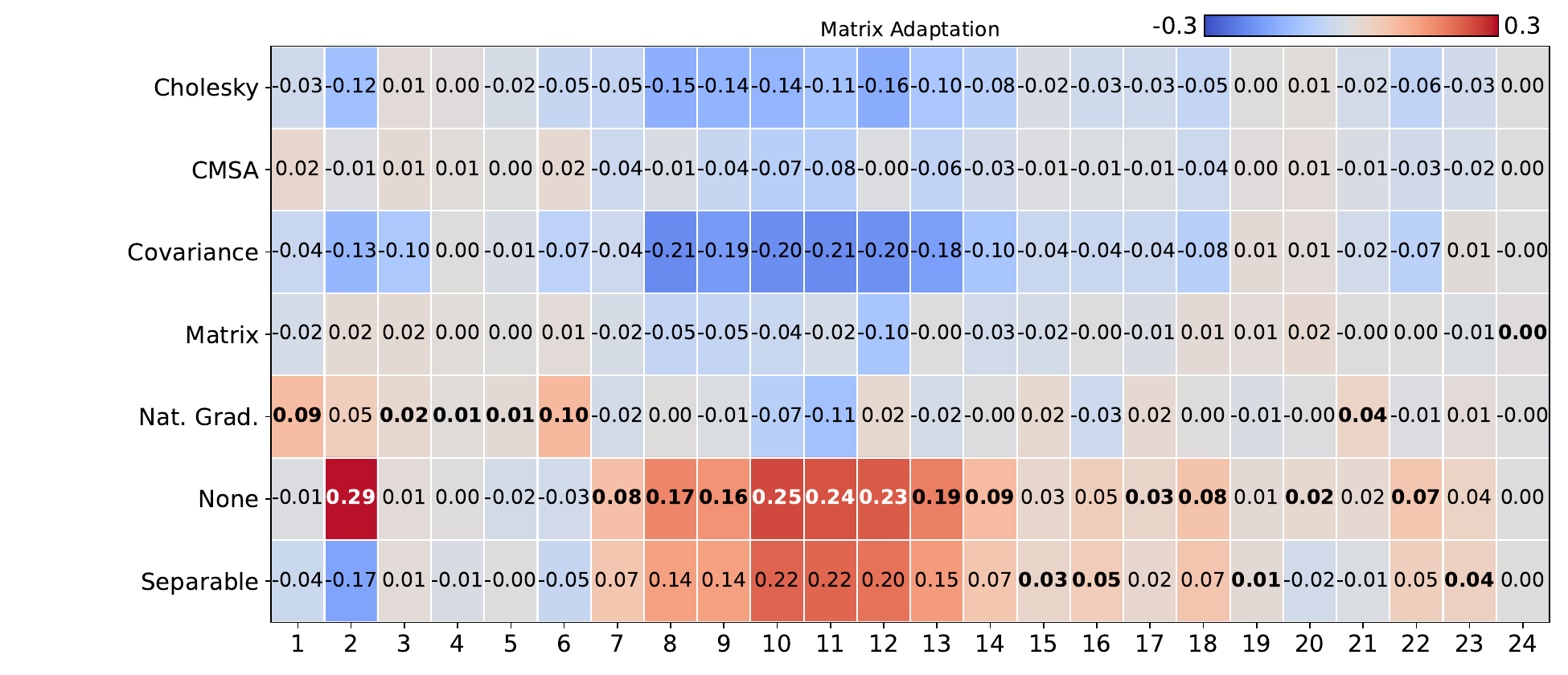}
\caption{
Per-function SHAP analysis for the \emph{Matrix Adaptation} module. Rows correspond to the available matrix adaptation strategies, and columns to the 24 BBOB functions. Each cell shows the median SHAP value over all evaluated configurations, indicating whether a given option tends to improve (negative values) or degrade (positive values) performance for that function.}
\label{fig:shap_matrix}
\end{figure}

\paragraph{Local analysis.}
To illustrate the insights provided by the SHAP analysis, we focus on the \emph{Matrix Adaptation} module, as this is also examined in detail in Section~\ref{modcma:sec:matrix:exp}. Similar results for the remaining modules are provided in Appendix~\ref{app:shap}. 
For this module, we analyze the \textsc{SHAP} values on a per-function basis, revealing how individual options contribute to performance across the BBOB suite. The results are visualized as a heatmap in Figure~\ref{fig:shap_matrix}, where each cell shows the median \textsc{SHAP} value for a given option and function. Immediately, the detrimental effect of disabling matrix adaptation on conditioned functions becomes apparent, particularly on the ellipsoid function $f_2$ and its rotated counterpart $f_{10}$. In contrast, separable adaptation yields a strong positive effect on $f_2$, while performing similarly poorly to no adaptation on $f_{10}$, consistent with expectations. More interestingly, on highly multimodal functions such as $f_4$, and $f_{24}$, the choice of matrix adaptation appears to have limited influence on performance. This suggests that success on these problems depends more on mechanisms for handling multimodality and escaping local optima than on adapting the sampling distribution. Overall, the Cholesky-based and standard covariance adaptation methods show the most consistently positive contributions across functions, which aligns with the trends observed in the controlled benchmarking results in Section~\ref{modcma:sec:matrix:exp}.

\section{Conclusion}
\label{sec:conclusion}
We presented \modcma, a modular framework for constructing and studying CMA-ES variants from interchangeable algorithmic components. By exposing modules for sampling, selection, step-size adaptation, matrix adaptation, and restarting within a unified implementation, \modcma enables the systematic combination, benchmarking, and configuration of a large space of CMA-ES variants.

Through a focused analysis of matrix adaptation, we showed that different adaptation mechanisms exhibit clear trade-offs between computational cost and optimization performance. In particular, alternatives such as Cholesky- and Matrix-based adaptation can substantially reduce runtime while retaining competitive performance on many benchmark problems. We further demonstrated the usefulness of the framework for automated configuration with \textsc{SMAC}~\cite{smac}, where per-function tuning on BBOB produced specialized variants that substantially improved over the default configuration and outperformed the established \textbf{bbob2009} reference set~\cite{hansen2010comparing} on half of all the functions. Finally, the \textsc{SHAP}-based analysis demonstrated that the resulting tuning data can be used to interpret the roles of individual modules and options, revealing both global trends and problem-specific effects.

Overall, these results show that \modcma is not only a practical implementation of many CMA-ES variants, but also a tool for studying algorithmic design choices and their interactions. Future work includes extending the framework with additional modules, improving support for mixed-integer and constrained optimization, and using the configuration space for dynamic or landscape-aware algorithm selection. Another promising direction is to combine modular CMA-ES variants with complementary optimization mechanisms, such as local search,  surrogate models, or constraint handling techniques~\cite{dufosse2021augmented}, to further improve performance in scenarios where standard CMA-ES variants are suboptimal.

\FloatBarrier

\bibliographystyle{plainnat}
\bibliography{biblio}
\newpage
\clearpage
\appendix
\section{Expected running time}\label{sec:ERT_full}

\begin{figure}[h]
    \centering
    \includegraphics[width=.95\textwidth]{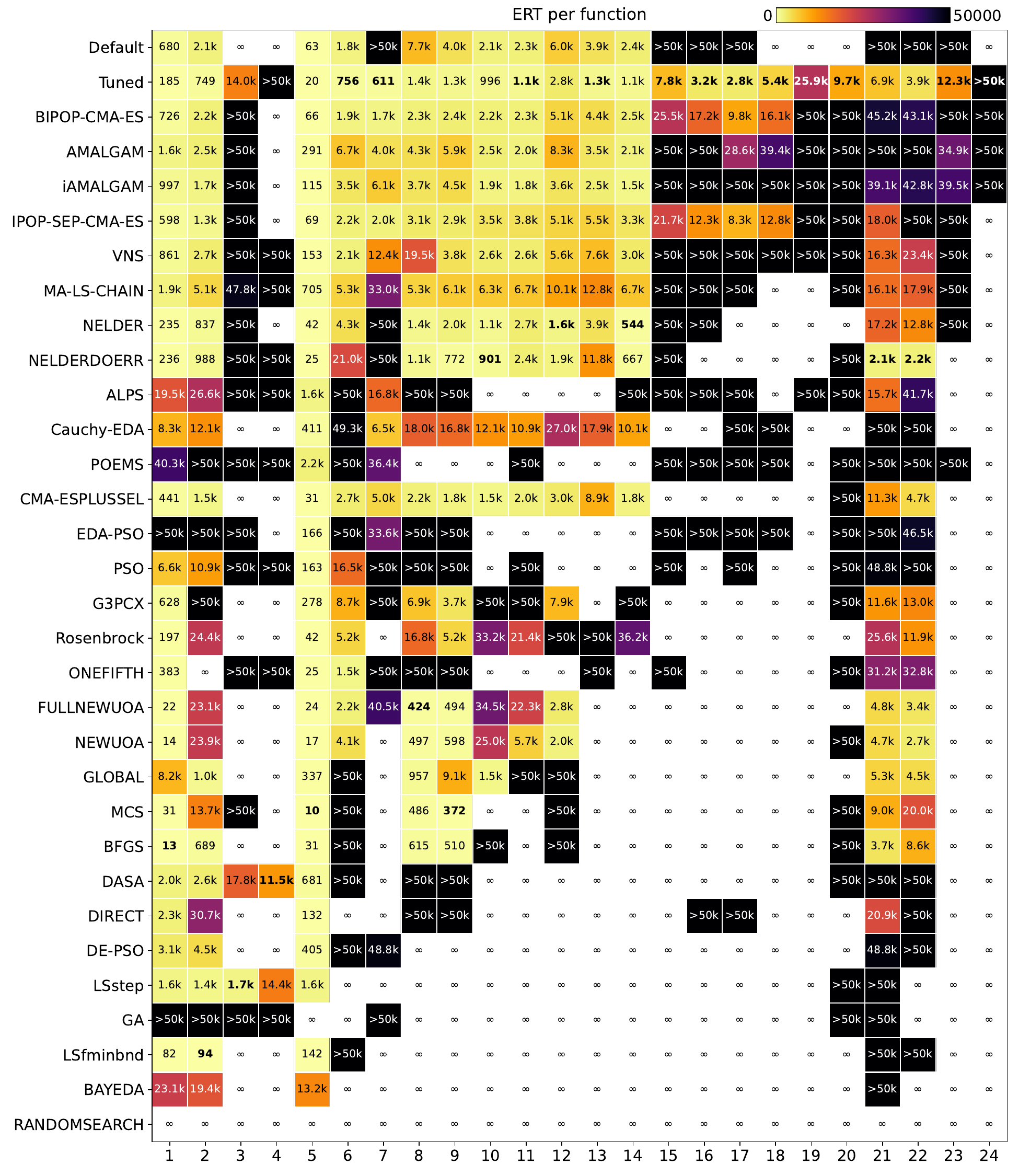}
    \caption{
    Expected running time (ERT) of the default and tuned \modcma configurations on the 24 BBOB functions in dimension $5$, together with the individual algorithms from the \textbf{bbob2009} benchmark portfolio. Lower values indicate better performance. The best (lowest) value per function is shown in bold. For some functions, algorithms fail to reach the target within the evaluation budget, resulting in large or infinite ERT values.
    }
        \label{fig:ERT_5D_full}
\end{figure}

\newpage

\section{Per Module SHAP contributions}
\label{app:shap}
\begin{figure}[h]
\centering
\includegraphics[width=\textwidth]{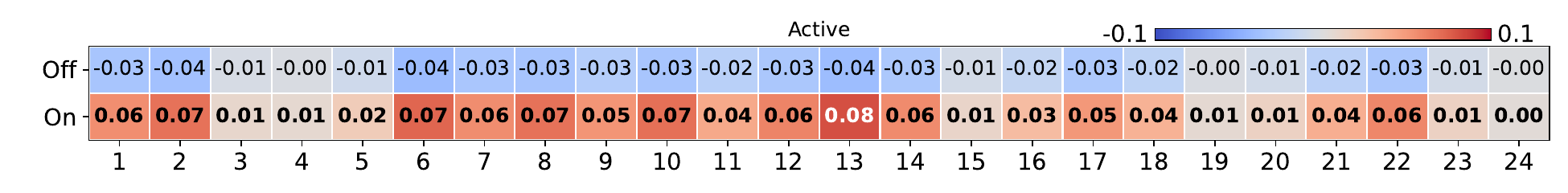}
\caption{Per-function SHAP analysis for the \emph{Active Update} module. Rows correspond to the available module options, and columns to the 24 BBOB functions. Each cell shows the median SHAP value over all evaluated configurations, indicating whether a given option tends to improve (negative values) or degrade (positive values) performance for that function.}
\label{fig:shap_active}
\end{figure}

\begin{figure}[h]
\centering
\includegraphics[width=\textwidth]{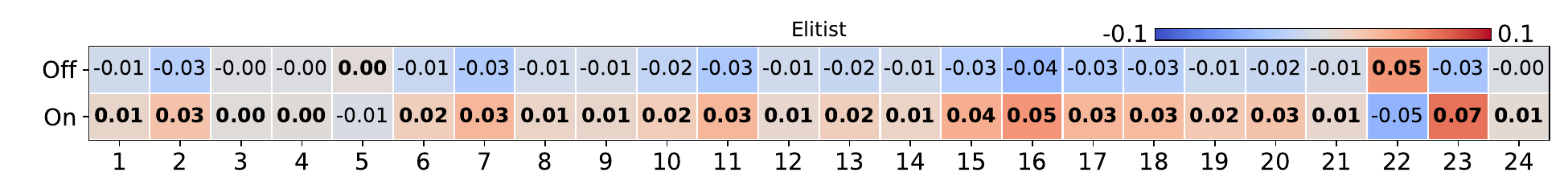}
\caption{Per-function SHAP analysis for the \emph{Elitism} module. Rows correspond to the available module options, and columns to the 24 BBOB functions. Each cell shows the median SHAP value over all evaluated configurations, indicating whether a given option tends to improve (negative values) or degrade (positive values) performance for that function.}
\label{fig:shap_elitist}
\end{figure}

\begin{figure}[h]
\centering
\includegraphics[width=\textwidth]{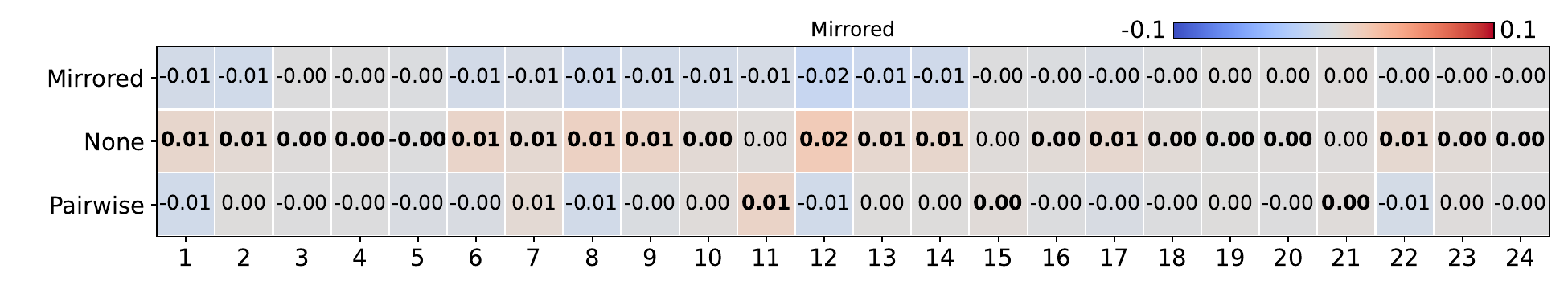}
\caption{Per-function SHAP analysis for the \emph{Mirrored Sampling} module. Rows correspond to the available module options, and columns to the 24 BBOB functions. Each cell shows the median SHAP value over all evaluated configurations, indicating whether a given option tends to improve (negative values) or degrade (positive values) performance for that function.}
\label{fig:shap_mirrored}
\end{figure}

\begin{figure}[h]
\centering
\includegraphics[width=\textwidth]{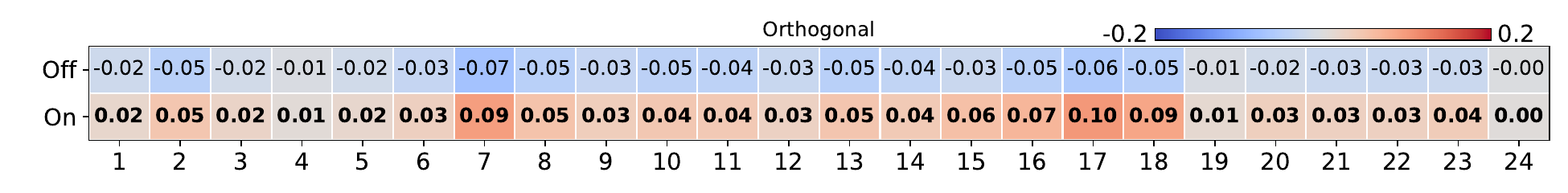}
\caption{Per-function SHAP analysis for the \emph{Orthogonal Sampling} module. Rows correspond to the available module options, and columns to the 24 BBOB functions. Each cell shows the median SHAP value over all evaluated configurations, indicating whether a given option tends to improve (negative values) or degrade (positive values) performance for that function.}
\label{fig:shap_orthogonal}
\end{figure}

\begin{figure}[h]
\centering
\includegraphics[width=\textwidth]{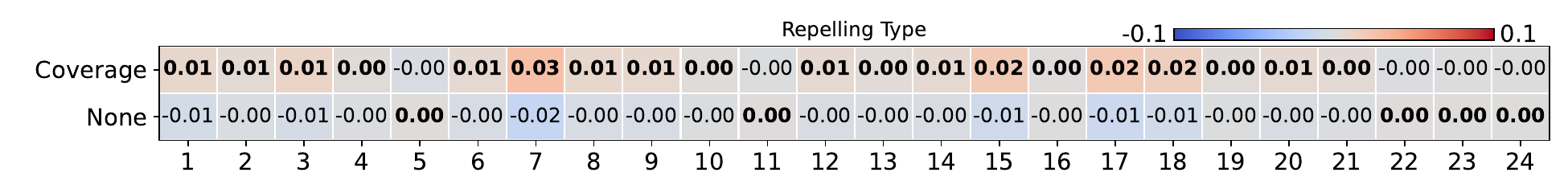}
\caption{Per-function SHAP analysis for the \emph{Repelling} module. Rows correspond to the available module options, and columns to the 24 BBOB functions. Each cell shows the median SHAP value over all evaluated configurations, indicating whether a given option tends to improve (negative values) or degrade (positive values) performance for that function.}
\label{fig:shap_repelling}
\end{figure}

\begin{figure}[h]
\centering
\includegraphics[width=\textwidth]{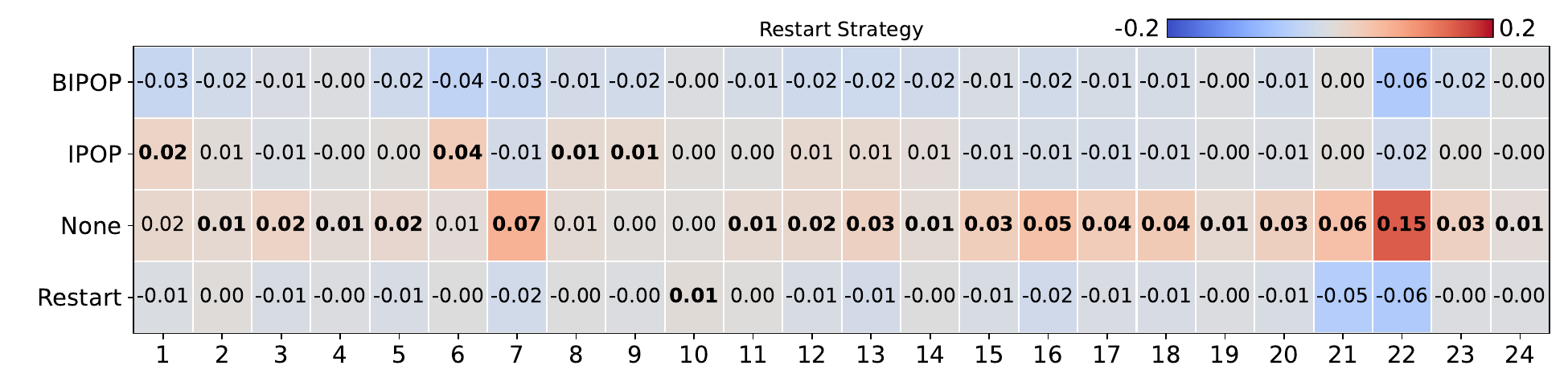}
\caption{Per-function SHAP analysis for the \emph{Restart Strategy} module. Rows correspond to the available module options, and columns to the 24 BBOB functions. Each cell shows the median SHAP value over all evaluated configurations, indicating whether a given option tends to improve (negative values) or degrade (positive values) performance for that function.}
\label{fig:shap_restart}
\end{figure}

\begin{figure}[h]
\centering
\includegraphics[width=\textwidth]{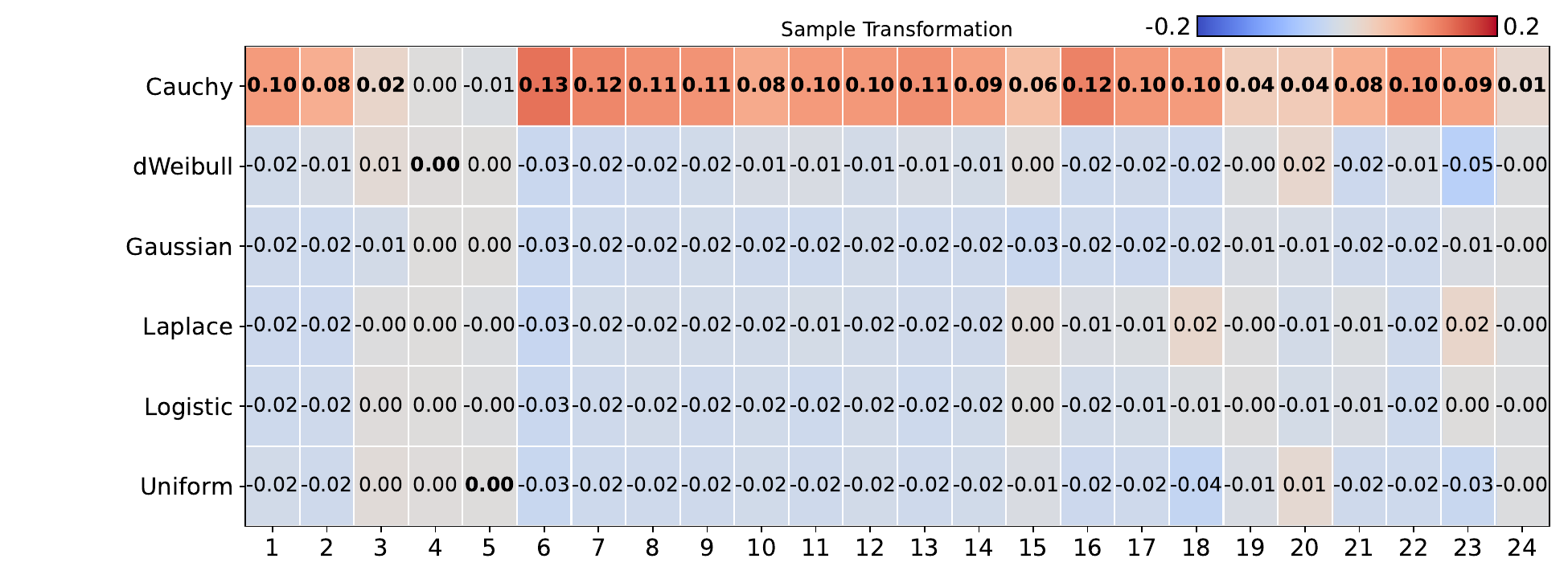}
\caption{Per-function SHAP analysis for the \emph{Sample Transformation} module. Rows correspond to the available module options, and columns to the 24 BBOB functions. Each cell shows the median SHAP value over all evaluated configurations, indicating whether a given option tends to improve (negative values) or degrade (positive values) performance for that function.}
\label{fig:shap_transformation}
\end{figure}

\begin{figure}[h]
\centering
\includegraphics[width=\textwidth]{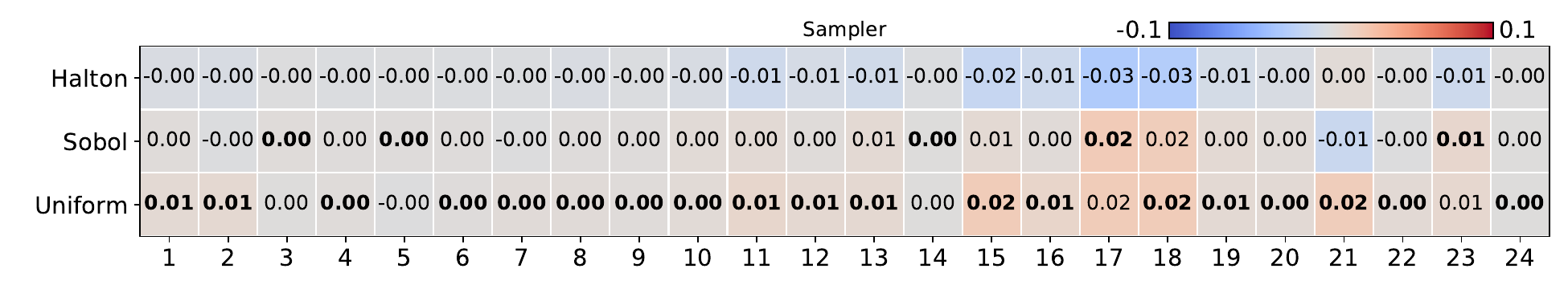}
\caption{Per-function SHAP analysis for the \emph{Base Sampler} module. Rows correspond to the available module options, and columns to the 24 BBOB functions. Each cell shows the median SHAP value over all evaluated configurations, indicating whether a given option tends to improve (negative values) or degrade (positive values) performance for that function.}
\label{fig:shap_sampler}
\end{figure}

\begin{figure}[h]
\centering
\includegraphics[width=\textwidth]{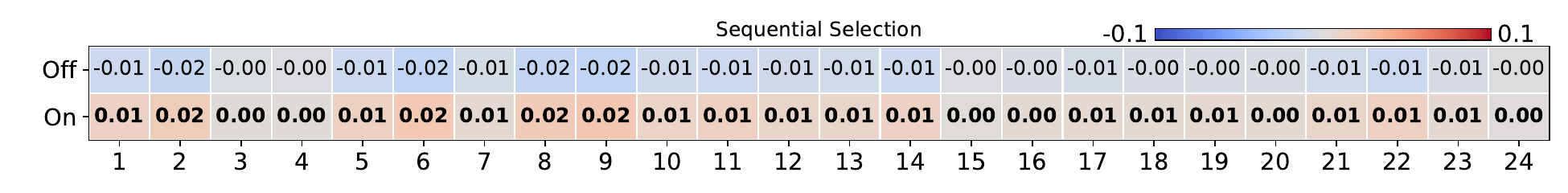}
\caption{Per-function SHAP analysis for the \emph{Sequential Selection} module. Rows correspond to the available module options, and columns to the 24 BBOB functions. Each cell shows the median SHAP value over all evaluated configurations, indicating whether a given option tends to improve (negative values) or degrade (positive values) performance for that function.}
\label{fig:shap_sequential}
\end{figure}

\begin{figure}[h]
\centering
\includegraphics[width=\textwidth]{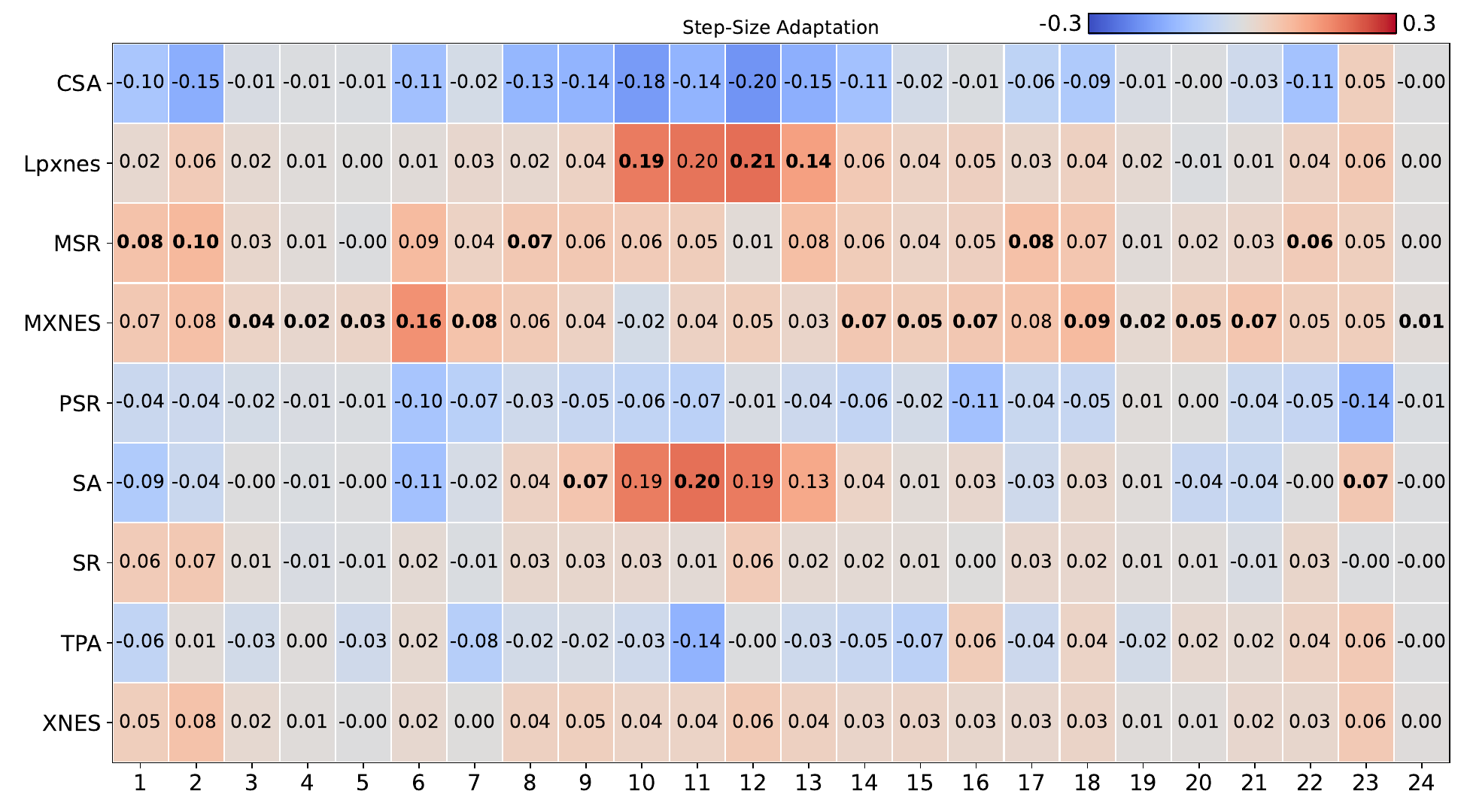}
\caption{Per-function SHAP analysis for the \emph{Step-Size Adaptation} module. Rows correspond to the available module options, and columns to the 24 BBOB functions. Each cell shows the median SHAP value over all evaluated configurations, indicating whether a given option tends to improve (negative values) or degrade (positive values) performance for that function.}
\label{fig:shap_ssa}
\end{figure}

\begin{figure}[h]
\centering
\includegraphics[width=\textwidth]{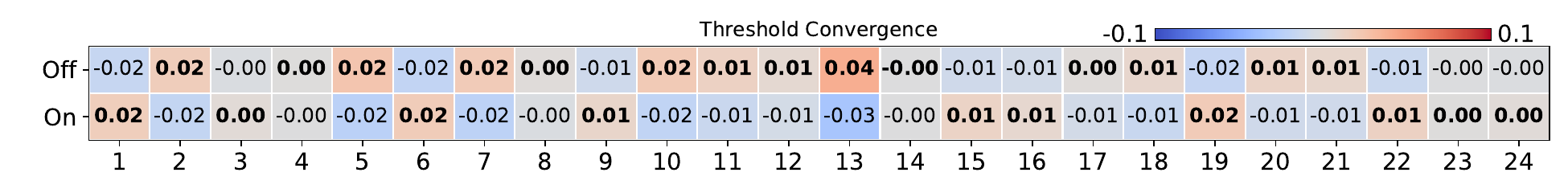}
\caption{Per-function SHAP analysis for the \emph{Threshold Convergence} module. Rows correspond to the available module options, and columns to the 24 BBOB functions. Each cell shows the median SHAP value over all evaluated configurations, indicating whether a given option tends to improve (negative values) or degrade (positive values) performance for that function.}
\label{fig:shap_threshold}
\end{figure}

\begin{figure}[h]
\centering
\includegraphics[width=\textwidth]{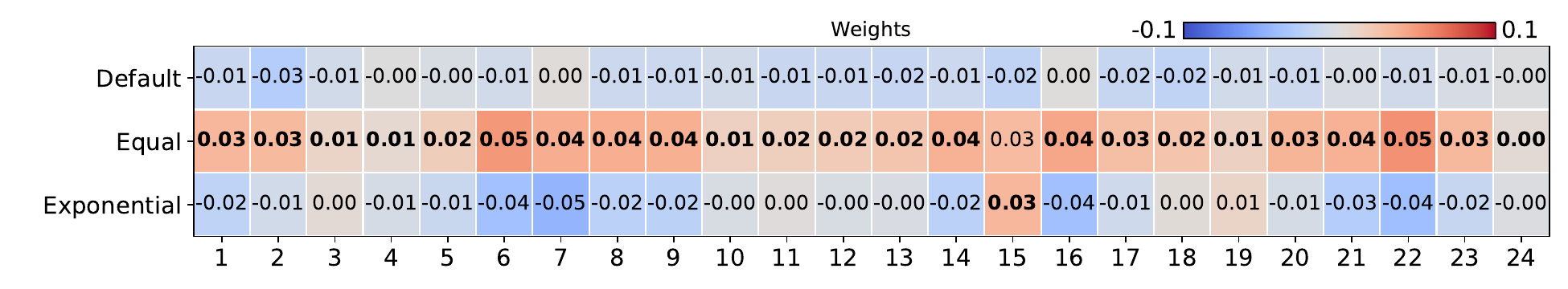}
\caption{Per-function SHAP analysis for the \emph{Recombination Weights} module. Rows correspond to the available module options, and columns to the 24 BBOB functions. Each cell shows the median SHAP value over all evaluated configurations, indicating whether a given option tends to improve (negative values) or degrade (positive values) performance for that function.}
\label{fig:shap_weights}
\end{figure}

\end{document}